\documentclass[manuscript, screen, nonacm]{acmart}
\AtBeginDocument{%
  }

\usepackage{tikz}
\usetikzlibrary{arrows.meta, positioning}

\usepackage{tcolorbox}
\tcbuselibrary{breakable}
\tcbuselibrary{skins}
\usepackage{xcolor}
\usepackage{fontawesome5}
\usepackage[table]{xcolor}
\usepackage{colortbl}
\usepackage{booktabs}
\usepackage{tabularx}
\usepackage{array}
\newcolumntype{Y}{>{\raggedright\arraybackslash}X}

\usepackage{graphicx}
\usepackage{xcolor}
\usepackage{tikz}
\usetikzlibrary{arrows.meta,calc}

\usepackage{fontspec}
\newfontfamily\EOsffamily{segoeui.ttf}[
  Path = fonts/,
  BoldFont = segoeuib.ttf,
  ItalicFont = segoeuii.ttf,
  BoldItalicFont = segoeuiz.ttf,
  Scale = MatchLowercase
]

\makeatletter
\def\@mkauthors{} 
\def\authorsaddresses{} 

\renewcommand{\@mktitle}{%
\begin{center}
{\LARGE\bfseries \@title \par}
\vspace{1.0em}

{\large
Muhammad Akhtar Munir$^{1}$ \quad
Muhammad Umer Sheikh$^{1}$ \quad
Akashah Shabbir$^{1}$ \\[0.25em]
Muhammad Haris Khan$^{1}$ \quad
Fahad Khan$^{1}$ \quad
Xiao Xiang Zhu$^{2}$ \\[0.25em]
Begüm Demir$^{3}$ \quad
Salman Khan$^{1}$ \\[0.75em]
}

{\normalsize
$^{1}$Mohamed bin Zayed University of Artificial Intelligence, Abu Dhabi, UAE \\[0.15em]
$^{2}$Technical University of Munich, Munich, Germany \\[0.15em]
$^{3}$BIFOLD and Technische Universität Berlin, Berlin, Germany
}

\vspace{0.8em}
\end{center}
}
\makeatother

\begin{document}

\hypersetup{
    citecolor=blue,
    linkcolor=blue,
    urlcolor=blue
}

\title{Agentic AI for Remote Sensing: Technical Challenges and Research Directions}

\author{Muhammad Akhtar Munir}
\email{akhtar.munir@mbzuai.ac.ae}
\affiliation{%
  \institution{Mohamed bin Zayed University of Artificial Intelligence}
  \city{Abu Dhabi}
  \country{UAE}
}

\renewcommand{\shortauthors}{Munir et al.}

\begin{abstract}
Earth Observation (EO) is moving beyond static prediction toward multi-step analytical workflows that require coordinated reasoning over data, tools, and geospatial state. While foundation models and vision-language models have advanced representation learning and language-grounded interaction in remote sensing, and agentic AI has shown strong potential for long-horizon reasoning and tool use, EO is not a straightforward extension of generic agentic AI. EO workflows operate on georeferenced, multi-modal, and temporally structured data, where operations such as reprojection, resampling, compositing, and aggregation transform the underlying state and can constrain later analysis. As a result, errors may propagate silently across steps, and correctness depends not only on internal coherence but also on geospatial consistency, temporally valid comparisons, and physical validity. This position paper argues that these challenges are structural rather than incidental. We examine the assumptions commonly made in generic agentic systems, analyze how they break in geospatial workflows, and characterize failure modes in multi-step EO pipelines. We then outline design principles for EO-native agents centered on structured geospatial state, tool-aware reasoning, verifier-guided execution, and validity-aware learning and evaluation. Building reliable geospatial agents, therefore, requires rethinking agent design around the physical, geospatial, and workflow constraints that govern EO analysis.
\end{abstract}



\maketitle

\section{Introduction}
\subsection{From Predictive Models to Decision-Oriented Models}

\definecolor{darkgold}{RGB}{184,134,11}
\begin{tcolorbox}[colback=gray!8, colframe=darkgold!75!black, boxrule=0.4pt, left=6pt, right=6pt, top=5pt, bottom=5pt]
{\large\color{darkgold!85!black}\faGlobe}\hspace{0.7em}%
\textbf{Earth Observation is not a single-step prediction problem.} It involves a sequence of decisions over data, tools, and geospatial state, demanding agentic models rather than standalone models.
\end{tcolorbox}

Over the past decade, remote sensing (RS) has been transformed by deep learning models built for well-defined EO tasks~\cite{ding2019learning, mou2019relation, ma2019deep, chen2019change, bandara2022changeformer, liu2024rotated}. These models substantially improved core applications such as land-cover and land-use classification, object detection in high-resolution imagery, semantic segmentation, and pixel-level change detection~\cite{chen2021remote, han2021align, wang2022unetformer}. In practice, this progress was driven by both stronger model families and stronger benchmarks: aerial detectors, encoder-decoder segmenters, and temporal change models matured alongside datasets such as DOTA, DIOR, NWPU VHR-10, xBD, FloodNet, and BigEarthNet~\cite{ding2019learning, ren2015faster, redmon2016you, wang2022unetformer, badrinarayanan2017segnet, chen2021remote, xia2018dota, li2020dior, cheng2016learning, gupta2019xbd, rahnemoonfar2021floodnet, clasen2025reben}. As a result, task-specific predictive models became the backbone of many operational remote sensing workflows. Yet their strength is also their limitation: they are designed for predefined tasks with fixed inputs and outputs, and therefore provide limited support for multi-step reasoning, adaptive tool use, or dynamically evolving analytical goals.

Building on task-specific models, the field has shifted toward EO foundation models that learn transferable geospatial representations from large unlabeled data. Early works such as S2MAE, CROMA, SatMAE, MMEarth, and OmniSat highlight the need to capture EO-specific structure, including spatial-spectral organization, cross-sensor alignment, and multimodal or semantically guided pretraining~\cite{li2024s2mae, fuller2023croma, wang2024multi, nedungadi2024mmearth, cong2022satmae, astruc2024omnisat}. This line of work has since expanded toward broader geospatial settings across modalities, resolutions, and sensors, including models such as Galileo, AnySat, Copernicus-FM, TerraFM, TerraMind, and Prithvi-2~\cite{tsenggalileo, astruc2025anysat, waldmann2025panopticon, han2024bridging, wang2025towards, danish2025terrafm, jakubik2025terramind, szwarcman2025prithvi, li2026fleximo, zhang2025skysense}. Parallel efforts emphasize temporal robustness, multi-scale learning, spectral and geometric biases, and efficient long-sequence modeling~\cite{feng2025tessera, smith2023earthpt, yao2023ringmo, reed2023scale, tang2023cross, noman2024rethinking, hong2024spectralgpt, li2024masked, duc2025satmamba, herzog2025olmoearth}, supported by increasingly large datasets and multimodal benchmarks~\cite{bastani2023satlaspretrain, bountos2025fomo, brown2025alphaearth}. Overall, this transition moves EO toward more general and transferable representations, but it remains primarily focused on representation learning rather than workflow-level reasoning.

In parallel, vision-language models introduced multimodal architectures that connect visual perception with natural language interaction and reasoning~\cite{li2023blip, li2024llava}, and were rapidly adapted to EO through models such as SkySenseGPT, GeoChat, SkyEyeGPT, LHRS-Bot, EarthGPT, and Falcon, enabling captioning, VQA, instruction following, and conversational interaction over satellite imagery~\cite{luo2024skysensegpt, kuckreja2024geochat, hu2025rsgpt, zhan2025skyeyegpt, bazi2024rs, muhtar2024lhrs, li2025lhrs, zhang2024earthgpt, yao2025falcon}. This paradigm broadened EO interaction from task-specific outputs to language-driven interpretation. At the same time, EO representation learning was strengthened through large-scale image-text alignment and CLIP-style transfer, alongside unified multimodal models that integrate heterogeneous inputs such as cross-sensor data and metadata~\cite{mall2023remote, Zhang_2024, liu2024remoteclip, wang2024skyscript, xiong2025dofa, shu2025earthmind, soni2025earthdial, irvinteochat, yuan2025omnigeo, feng2025urbanllava}. Fine-grained capabilities also advanced from region-level grounding to pixel-level reasoning and unified grounding-segmentation pipelines~\cite{zhang2024earthmarker, shabbir2025geopixel, ou2025geopix, wang2024ringmogpt, yao2025remotesam, zhu2025skysense, shu2026terrascope, li2025segearth}, supported by benchmarks for VQA, VLM evaluation, and ultra-high-resolution understanding~\cite{luo2025large, danish2025geobench, wang2025xlrs}. Despite these advances, EO VLMs still largely treat analysis as input-output prediction rather than sequential, tool-grounded reasoning over geospatial state.

Alongside advances in foundation models and EO vision-language models, agentic AI has emerged as a broader framework for problems that require sequential reasoning, adaptive decision making, and interaction with external tools. 
Rather than producing a single prediction, agentic models operate over trajectories of actions and observations: they decompose goals, select tools, execute intermediate steps, and use the resulting outputs to guide subsequent reasoning. 
Across the broader agent literature, frameworks such as ToolOrchestra and VerlTool study tool orchestration and reinforcement-learning-based tool use, while OpenThinkIMG, SpaceTools, OctoTools, and DeepEyes explore visual tool calling, multimodal action policies, and long-horizon perception-action loops~\cite{su2025toolorchestra, jiang2025verltool, su2025openthinkimg, chen2025spacetools, wu2025tool, lu2025octotools, zheng2025deepeyes}. Collectively, these works establish core ingredients of agentic AI, including tool use, sequential control, and trajectory-level optimization, but typically in settings where tools are relatively abstract and domain constraints are limited.

\begin{figure*}[t]
  \centering
  {%
    \let\sffamily\EOsffamily
    \let\EOUIFont\EOsffamily
    \input{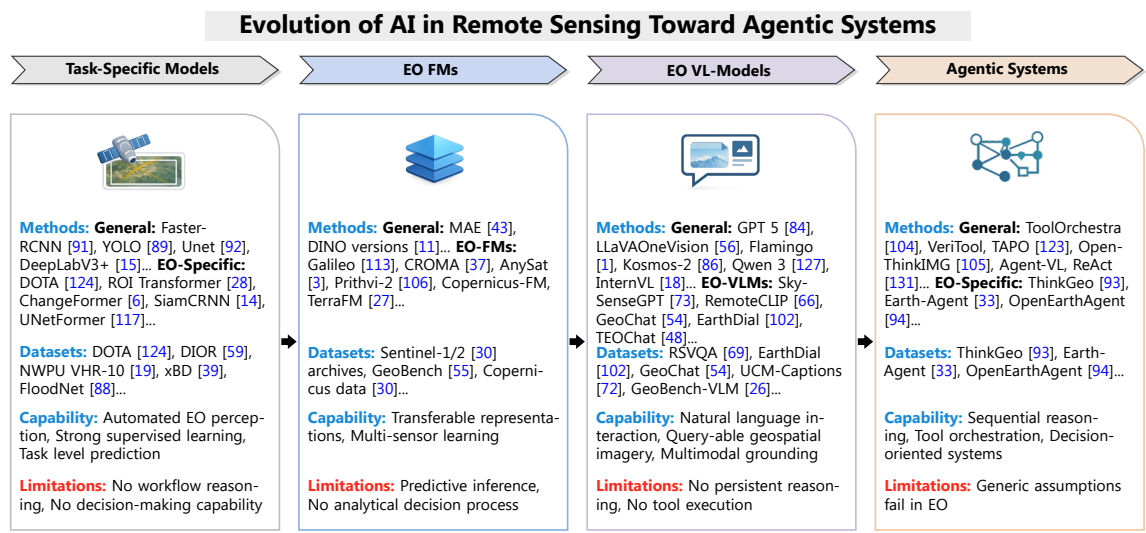}%
  }\vspace{-1.5em}
  \caption{Evolution of AI paradigms in remote sensing toward agentic models. The field has progressed from task-specific predictive models to EO foundation models, EO vision-language models, and emerging agentic frameworks. Each stage expands the capability space from automated perception and transferable representation learning to language-grounded interaction and sequential tool use. At the same time, this progression exposes a mismatch between generic agentic assumptions and EO workflow constraints, motivating agentic models that reason over geospatial state, tool effects, and validity rather than final predictions alone. \textit{Drawings from ChatGPT.}}
  \label{fig:eoevolution}
\end{figure*}

Within remote sensing and geospatial analysis, a first wave of work has begun to translate these ideas into more realistic EO settings. Some efforts focus on benchmark and environment design, showing that geospatial agents must operate over realistic APIs, map-based interfaces, and long-horizon user commands rather than simplified question-answer templates, as in GeoLLM-QA, GeoLLM-Engine, and UnivEARTH~\cite{singh2024evaluating, singh2024geollm, kao2025towards}. Others develop geospatial copilots and agentic frameworks that combine tool orchestration with retrieval, state-driven workflows, map interaction, and domain knowledge, including RS-Agent, MapAgent, ThinkGeo, OpenEarthAgent, Earth-Agent, and REMSA~\cite{stamoulis2025geo, xu2024rs, hasan2025mapagent, shabbir2025thinkgeo, shabbir2026openearthagent, feng2025earth, chen2025remsa}. A further set of models targets domain-specialized interactive analysis, including comprehensive change interpretation and geologic map understanding through coordinated perception, reasoning, and external knowledge, as in Change-Agent and PEACE~\cite{liu2024change, huang2025peace}. Collectively, these works mark an important transition from predictive EO models toward workflow-aware geospatial reasoning. At the same time, they remain early steps: current models are still limited in coverage, robustness, and validation, and they expose how far EO agents remain from geo-valid and scientifically accountable long-horizon analysis. Figure~\ref{fig:eoevolution} summarizes this progression from task-specific predictive models to foundation models, vision-language models, and emerging agentic frameworks, while also highlighting where the mismatch between generic agentic assumptions and EO workflow realities begins to emerge.

\begin{tcolorbox}[colback=gray!8, colframe=darkgold!75!black, boxrule=0.4pt, left=6pt, right=6pt, top=5pt, bottom=5pt]
{\large\color{darkgold!85!black}\faGlobe}\hspace{0.7em}%
\textbf{An EO agent is a geospatial state updater.} It selects parameterized tools, modifies structured EO state, and must preserve geospatial and physical validity across intermediate transitions.
\end{tcolorbox}

\subsection{Why Agentic AI Requires Reconsideration in Remote Sensing}

At first glance, agentic AI appears well aligned with EO workflows: both involve multi-step reasoning, tool use, and sequential decision-making~\cite{jiang2025verltool, wu2025tool, su2025openthinkimg, chen2025spacetools}. 
However, many current agentic frameworks and benchmarks are developed in environments where tools behave like predictable functions, actions are largely recoverable, and errors can often be detected before they substantially affect later steps~\cite{su2025toolorchestra, liu2025agent0, su2025openthinkimg}. Under these assumptions, supervision and reward signals remain relatively stable because intermediate mistakes are often easier to identify or correct.

\begin{figure*}[t]
  \centering
  \includegraphics[width=\textwidth]{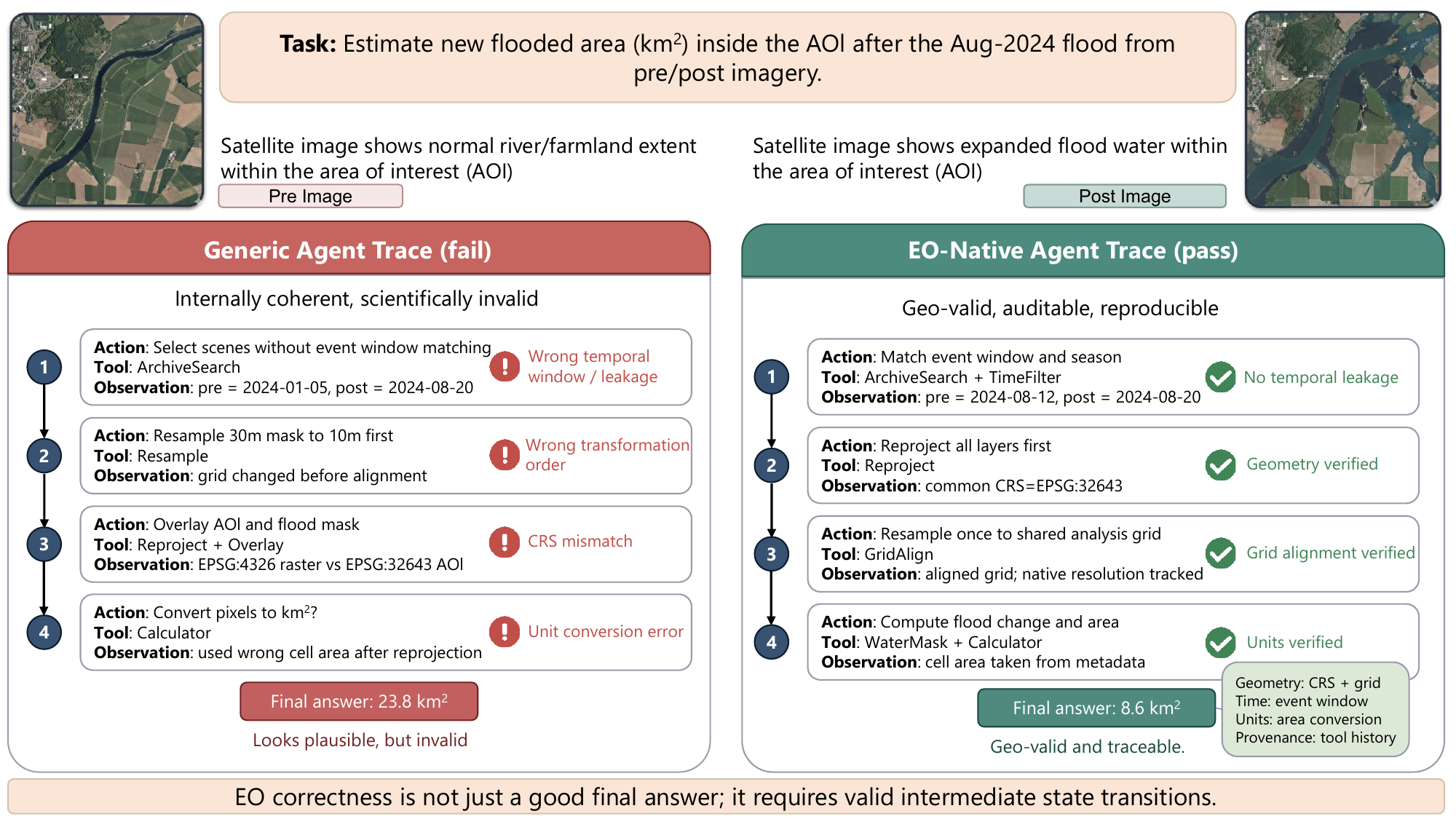}
  \caption{Illustrative comparison of generic and EO-native agent traces for flood-area estimation from pre/post imagery. The generic trace reaches a plausible-looking but invalid result because the workflow contains representative EO failure modes, including temporal-window mismatch, incorrect transformation order, coordinate reference system (CRS) mismatch, and unit-conversion error. The EO-native trace illustrates the corresponding valid workflow: matching the event window, reprojecting layers to a common CRS, aligning data to a shared analysis grid, and computing area using metadata. The example highlights that EO correctness depends not only on the final answer, but also on valid intermediate state transitions. \textit{Figure developed with the help of ChatGPT.}}
  \label{fig:eo_trace_compare}
\end{figure*}

EO workflows break these assumptions in important ways. They operate over georeferenced, multi-modal, and temporally structured data, and they rely on transformations such as reprojection, resampling, spatial aggregation, and index computation whose effects are stateful and often order-dependent. These operations do not merely retrieve information; they modify the data that later steps must reason over. As a result, an early mistake in spatial alignment, temporal selection, or preprocessing may silently propagate through the rest of the pipeline and only become visible at the final stage of analysis. Figure~\ref{fig:eo_trace_compare} illustrates this point through a representative contrast: a generic agent may produce a plausible final answer through scientifically invalid intermediate steps, whereas an EO-native agent enforces geo-valid and reproducible state transitions throughout the workflow.

A related issue concerns how correctness is defined. In many agentic settings, optimization can rely on internal signals such as task completion, self-consistency, or efficiency-based rewards~\cite{liu2025agent0, thawakar2025evolmm, zhai2025agentevolver}. In EO, however, correctness is not just an internal property of the reasoning trace. It depends on externally grounded criteria such as physical validity, geospatial consistency, and agreement with observed phenomena across space and time. Even high-level choices such as sensing modality or model selection may already require reasoning about scale, coverage, and data suitability rather than generic task-level optimization~\cite{chen2025remsa}.
These differences suggest that agentic AI for remote sensing should not be viewed as a straightforward extension of existing tool-use frameworks. The challenge is not simply to attach geospatial tools to a generic agent, but to rethink agent design, learning objectives, and evaluation around structured geospatial state and the scientifically constrained nature of EO workflows.

\begin{tcolorbox}[colback=gray!8, colframe=darkgold!75!black, boxrule=0.4pt, left=6pt, right=6pt, top=5pt, bottom=5pt]
{\large\color{darkgold!85!black}\faGlobe}\hspace{0.7em}%
\textbf{Generic agentic AI cannot be transferred to EO.} EO workflows are stateful, externally validated, and vulnerable to silent error propagation across preprocessing, analysis, and interpretation steps.
\end{tcolorbox}

\subsection{Scope and Contributions of This Position Paper}

The discussion above points to a central claim of this paper: EO is not simply another application domain for generic agentic AI. Although recent agentic models are effective at long-horizon reasoning and tool interaction, they are typically developed for settings in which tools behave as abstract operations, intermediate states are relatively easy to observe, and errors can often be detected or corrected internally. EO workflows differ in more fundamental ways. They operate on geospatially referenced, physically constrained, and often multi-modal data, where intermediate transformations are state-changing, can be order-dependent, and may be partly irreversible, while early mistakes can silently affect downstream analysis. These differences motivate a dedicated examination of agentic AI through the lens of remote sensing.

At the same time, our argument is deliberately scoped. Recent papers have already begun to survey agentic AI in remote sensing, including work that organizes the area around foundations, agent types, enabling models, and evaluation, as well as broader survey treatments covering formal definitions, taxonomies, applications, datasets, and benchmarks~\cite{talemi2026agentic,tangintelligent}. Our paper is complementary but different in purpose. We do not aim to provide a comprehensive survey of models, an exhaustive taxonomy of architectures, or a broad inventory of applications and benchmarks. 
Instead, we make a more targeted claim: \emph{remote sensing is a distinct agentic domain where core assumptions of generic agentic AI do not strictly hold.}
EO reasoning operates over a structured geospatial state rather than an abstract sequence of tool calls. Many transformations are order-dependent and may be partly irreversible, so early mistakes can silently propagate through later stages of the workflow. Correctness, therefore, cannot be judged by internal coherence alone; it must be grounded in external criteria such as spatial consistency, temporal and physical validity. What distinguishes this paper is thus not coverage of the field, but a sharper conceptual argument: reliable agentic EO requires rethinking agent design, learning objectives, and evaluation around geospatial state \& validity, and trajectory-level accountability.

Accordingly, this paper does not introduce a new benchmark, propose a single agent architecture, or attempt a conventional survey of the agentic AI literature. Rather, it makes three focused contributions:
\begin{itemize}
    \item \textbf{Positioning EO as a distinct agentic domain.}  
    We show that remote sensing is not a straightforward application of generic tool-using agents by making explicit the assumptions such models usually rely on and explaining why they become fragile in EO workflows.

    \item \textbf{Identifying EO-specific failure modes.}  
    We analyze how these assumptions break in geospatial pipelines, leading to invalid tool ordering, spatial or temporal misalignment, silent propagation of preprocessing errors, and outputs that appear plausible but violate geospatial or physical validity.

    \item \textbf{Deriving EO-native design and evaluation principles.}  
    We derive principles for EO-native agents based on structured geospatial state, state-transforming scientific tools, verifier-guided reasoning, validity-aligned learning objectives, and trajectory-level evaluation beyond final-answer accuracy.
\end{itemize}
The rest of the paper is organized as follows. Section~\ref{sec2} reviews recurring paradigms and assumptions in current agentic AI models. Section~\ref{sec3} explains why EO constitutes a distinct agentic domain with structural properties that challenge those assumptions. Section~\ref{sec4} analyzes the resulting limitations and failure modes of current agentic methods in remote sensing. Section~\ref{sec5} outlines design principles for EO-native agentic models, Section~\ref{sec6} discusses research directions in benchmarking, learning, and evaluation, and Section~\ref{sec7} concludes with a broader outlook on agentic EO models as decision-support instruments.

\section{Agentic AI: Core Paradigms and Assumptions}
\label{sec2}
\subsection{Agentic Architectures}

Recent agentic AI models typically follow a small number of recurring architectural patterns that combine learned perception with explicit decision-making mechanisms. A common design uses tool-augmented language or vision-language models, where a base model generates intermediate reasoning steps and invokes external tools through structured function calls. Another widely adopted pattern is the planner-executor-verifier pipeline, in which task decomposition, tool execution, and result validation are handled by distinct agent roles. Many models further use supervised fine-tuning over multi-step reasoning trajectories to stabilize tool usage and structured decision-making, sometimes followed by reinforcement learning under task-specific objectives.
Across these architectures, reasoning and action are interleaved. Instead of producing a single output in one inference step, agents operate over sequences of decisions: they decompose tasks, select tools dynamically, execute operations, and use intermediate outputs to guide subsequent reasoning steps. This paradigm has been explored in recent tool orchestration frameworks and tool-integrated reinforcement learning approaches~\cite{wu2025tool,su2025toolorchestra,jiang2025verltool,su2025openthinkimg,liu2025agent0,hasan2025mapagent,zhai2025agentevolver}. In these works, tool selection and ordering are treated as part of the learned policy, enabling more flexible problem-solving than fixed analytical pipelines.

\begin{figure}[t] 
  \centering
  \includegraphics[width=\columnwidth]{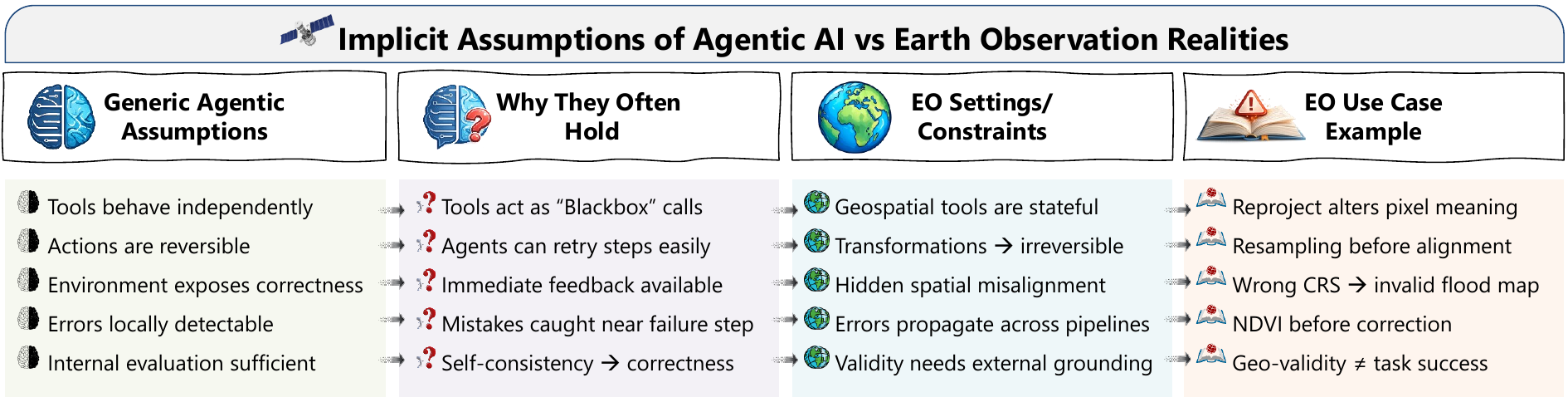}
  \caption{Implicit assumptions underlying generic agentic AI models and their mismatch with EO workflows. The figure summarizes a compact set of recurring assumptions often made in generic tool-use settings, e.g., independent tools, reversible actions, exposed correctness signals, locally detectable errors, and sufficient internal evaluation. These assumptions are often reasonable in conventional settings but become fragile in EO workflows, where geospatial operations are stateful, transformations may be irreversible, spatial misalignment can remain hidden, and validity requires externally grounded verification. Concrete EO use cases illustrate how seemingly correct agent behavior can still lead to scientifically invalid outcomes, motivating EO-specific agentic design principles. \textit{Drawings from ChatGPT.}}
  \label{fig:eoassumption}
\end{figure}

However, many current agentic frameworks implicitly assume that tool actions can be revised or corrected through subsequent decisions. In EO workflows, this assumption becomes fragile. Many geospatial operations directly transform the data used in later steps, and these transformations may be partly irreversible. Once operations such as resampling, reprojection, or spatial aggregation are applied, the original signal may be altered or partially lost. Incorrect actions can therefore propagate silently through later stages of the workflow without producing immediate feedback. This weakens observable learning signals and complicates credit assignment, limiting the effectiveness of standard agentic learning strategies when applied to geospatial analytical pipelines. Within geospatial settings, recent models have begun to explore retrieval-guided orchestration, state-driven prompting, and specialized interactive analysis pipelines, as in RS-Agent~\cite{xu2024rs}, Geo-OLM~\cite{stamoulis2025geo}, Change-Agent~\cite{liu2024change}, PEACE~\cite{huang2025peace}, and OpenEarthAgent~\cite{shabbir2026openearthagent}. These recurring assumptions and their mismatch with EO workflows are illustrated in Figure~\ref{fig:eoassumption}.

\begin{tcolorbox}[colback=gray!8, colframe=darkgold!75!black, boxrule=0.4pt, left=6pt, right=6pt, top=5pt, bottom=5pt]
{\large\color{darkgold!85!black}\faGlobe}\hspace{0.7em}%
\textbf{Agentic AI assumes tools can be selected, revised, and evaluated through feedback.} In EO, these assumptions weaken because tool calls can alter the geospatial state and constrain analysis.
\end{tcolorbox}

\subsection{Implicit Design Assumptions}

Despite differences in architecture and training strategy, many agentic AI models rely on a common set of implicit design assumptions that simplify both learning and evaluation. These assumptions are often reasonable in domains such as web interaction, code execution, or simulated environments, but they are rarely stated explicitly. Figure~\ref{fig:eoassumption} summarizes them in compact form. The most recurrent ones include:

\begin{itemize}

\item \textbf{Tools behave independently.}  
Tools are often modeled as largely independent functions whose effects depend primarily on current inputs rather than accumulated workflow state.

\item \textbf{Actions are reversible.}  
Incorrect actions are often treated as recoverable or limited in side effects, allowing agents to retry or revise decisions without strongly constraining later reasoning.

\item \textbf{The environment exposes correctness.}  
The environment is assumed to provide enough feedback for the agent to judge whether an action is correct, with relatively few hidden effects that appear only later.

\item \textbf{Errors are locally detectable.}  
Mistakes are often assumed to become visible near the step that produced them, rather than propagating silently across longer reasoning chains.

\item \textbf{Internal evaluation is sufficient.}  
Task success is often judged through internal signals such as logical consistency, task completion indicators, learned evaluators, or reward proxies assumed to track task quality closely enough for optimization to improve behavior.

\end{itemize}

These assumptions simplify the design and training of agentic models, particularly when using supervised fine-tuning or reinforcement learning over action trajectories. At the same time, they constrain the environments in which such models can operate reliably. As later sections will show, EO workflows violate several of these assumptions because geospatial data processing is structured, state-dependent, and often only partially observable.

\subsection{Progress in Tool-Integrated and RL-Based Agents}

Recent work shows that agents can learn to coordinate tool use, manage multi-step workflows, and optimize computational efficiency through reinforcement learning~\cite{su2025toolorchestra,jiang2025verltool,wu2025tool,su2025openthinkimg,lu2025scaling}. In these models, tool selection and sequencing are treated as part of the agent's policy, enabling end-to-end optimization over reasoning and action trajectories.
This is an important step toward more flexible agentic models, but most current frameworks remain domain-agnostic in how they represent tools and their effects. Tools are typically treated as black-box operators with fixed behavior, and learning focuses primarily on when to call a tool, which tool to call, and in what order. In EO, however, tools correspond to transformations over geospatial data, and their outputs must satisfy constraints related to spatial reference systems, resolution, temporal alignment, and physical validity. Optimizing orchestration alone is therefore insufficient to guarantee correct EO analytical workflows.

\subsection{Self-Improving and Self-Evolving Agents}

Self-improving and self-evolving agents aim to reduce dependence on explicit human supervision by generating tasks, critiques, or reward signals internally. These mechanisms can improve performance in visual and multimodal reasoning when external supervision is limited~\cite{liu2025agent0,thawakar2025evolmm,zhai2025agentevolver}. Their usefulness, however, depends on the assumption that correctness can be inferred from internal signals such as agreement across reasoning paths, consistency among agent roles, or outputs of learned critics.

In EO, internal coherence is not sufficient to guarantee valid analysis. Correctness depends on physical validity, geospatial consistency, and agreement with observed phenomena across space and time. An agent may therefore produce outputs that appear logically consistent while still violating sensor characteristics, spatial reference systems, or physical constraints. Under such conditions, self-improving agents may converge toward stable but incorrect analytical strategies, consistent with recent findings that internally generated training signals can reinforce misaligned behavior~\cite{shao2025your}. Self-improvement in EO, therefore, requires external validation signals and domain-aware constraints that enforce geospatial validity throughout the analytical process.

\section{What Makes Remote Sensing a Distinct Agentic Domain}
\label{sec3}

\begin{figure}[t] 
  \centering
  \includegraphics[width=\columnwidth]{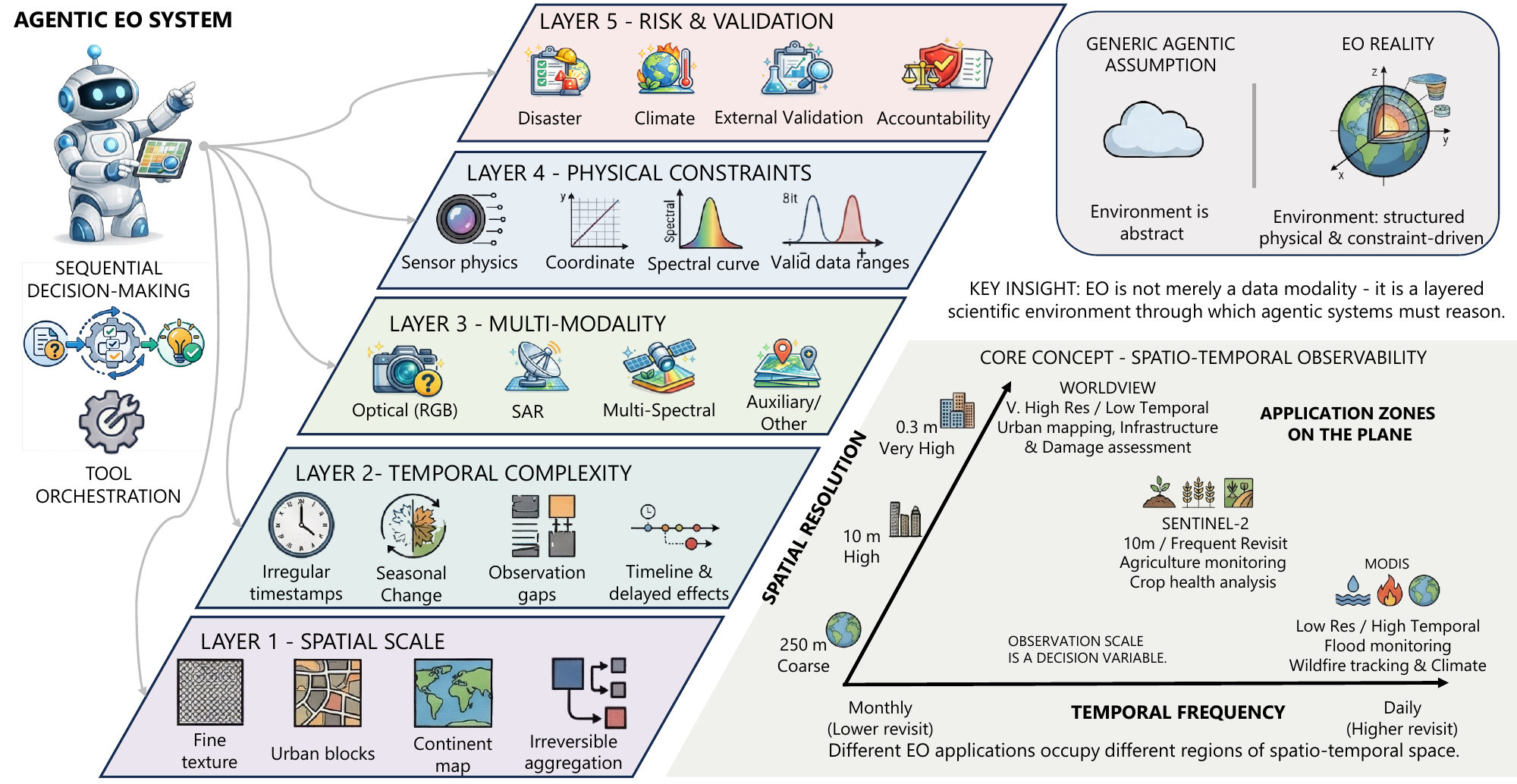}
  \caption{Structural properties of Earth observation environments for agentic models. EO reasoning operates within a layered environment characterized by spatial scale, temporal complexity, multi-modality, physical constraints, and risk-aware validation. These layers jointly define the conditions under which agentic EO models must perform sequential decision-making and tool orchestration. The figure also illustrates spatio-temporal observability, where different EO applications occupy distinct regions of the spatial-resolution and revisit-frequency space, making observation scale a critical decision variable for agentic reasoning. The sensing sources shown are representative rather than exhaustive. \textit{Figure developed with the help of ChatGPT.}}
  \label{fig:eostructure}
\end{figure}

Remote sensing exhibits structural characteristics that are intrinsic to EO data and analytical workflows. These characteristics distinguish EO from the environments in which many current agentic AI models have been designed and evaluated. As a result, several assumptions described in Section~\ref{sec2}, particularly those concerning scale, object definition, tool abstraction, temporal reasoning, and correctness evaluation, become difficult to sustain in geospatial analysis. Understanding these properties is therefore essential for designing agentic models that operate reliably in EO workflows. Figure~\ref{fig:eostructure} organizes them as layered constraints of the EO environment, spanning spatial scale, temporal complexity, sensing modality, physical constraints, and risk-aware validation.

\begin{tcolorbox}[colback=gray!8, colframe=darkgold!75!black, boxrule=0.4pt, left=6pt, right=6pt, top=5pt, bottom=5pt]
{\large\color{darkgold!85!black}\faGlobe}\hspace{0.7em}%
\textbf{Earth Observation is a distinct agentic domain.} Its workflows are shaped by scale, context, modality, time, and physical constraints, which jointly govern how data, tools, and reasoning interact.
\end{tcolorbox}

\subsection{Spatial Scale, Resolution, and Global Coverage}

EO data span an unusually wide range of spatial scales, from localized observations covering a few square kilometers to continuous planetary-scale coverage~\cite{francis2024major, xia2018dota, christie2018functional}. 
Observations are acquired at heterogeneous spatial resolutions and revisit intervals determined by sensor design, orbital parameters, and acquisition policies~\cite{CopernicusSentinels}. 
Consequently, scale selection is not simply a preprocessing choice but a central modeling decision: it determines which physical phenomena are observable, how signals aggregate or fragment, and which interpretations remain scientifically valid~\cite{reed2023scale}.  

For agentic models, this introduces a decision dimension that is largely absent in conventional image reasoning tasks. Choices about spatial resolution, tiling strategy, aggregation level, and geographic coverage are often tightly coupled with the analytical objective, and many of these transformations are difficult to reverse once applied. Treating scale as implicit or fixed, a common assumption in existing agentic pipelines, can therefore produce analyses that appear internally coherent yet lack geospatial validity. Agentic EO models must instead reason explicitly about spatial scale and geographic coverage as part of the analytical process. Related EO pretraining and multimodal work similarly treat scale as a first-class variable~\cite{tang2023cross, reed2023scale, li2026fleximo, jiang2405lemevit, zhang2025skysense, brown2025alphaearth, wang2025xlrs}.

\subsection{Object Ambiguity and Context Dependence}

Objects of interest in remote sensing frequently lack clear or stable visual boundaries. 
Their appearance can vary significantly with spatial resolution, sensing modality, illumination conditions, and acquisition geometry. 
Moreover, many EO targets are defined contextually rather than visually. 
Examples include informal settlements, flooded regions, burned areas, or damaged infrastructure, whose identification often requires temporal comparison or auxiliary information rather than a single image observation~\cite{rahnemoonfar2021floodnet, gupta2019xbd}.  

This ambiguity challenges the assumption that objects can be consistently localized and manipulated through purely visual representations. In EO workflows, object identity is often determined by spatial context, temporal evolution, and domain-specific definitions. Agentic models must therefore integrate contextual reasoning and auxiliary data sources rather than relying solely on visual abstractions such as bounding boxes or segmentation masks. Related work on spatial-language grounding and text-guided geolocalization, such as GeoText-1652~\cite{chu2024towards}, further highlights the importance of relational spatial language beyond isolated object recognition.

\subsection{Multi-Modal, Physics-Driven Measurements}

Remote sensing relies on a diverse set of sensing modalities, including panchromatic/RGB optical imagery, multispectral and hyperspectral optical measurements, synthetic aperture radar (SAR), and meteorological reanalysis data.
Each modality captures different physical interactions between the Earth's surface, atmosphere, and the sensing system, and therefore exhibits distinct noise characteristics, uncertainties, and preprocessing requirements~\cite{fuller2023croma, wang2025towards, szwarcman2025prithvi, nedungadi2024mmearth, guo2024skysense}. This diversity is also reflected in recent EO foundation and multimodal models designed for heterogeneous sensors and cross-modal fusion~\cite{soni2025earthdial, jakubik2025terramind, astruc2024omnisat, jain2022self, bountos2025fomo, zhang2024earthgpt, xiong2025dofa, yuan2025omnigeo}.

For example, SAR observations encode surface roughness and moisture through radar backscatter, whereas optical sensors measure reflected radiation influenced by illumination conditions and atmospheric effects. 
Agentic models must therefore reason not only about which tools to apply, but also about which sensing modality is appropriate for a given analytical objective and how modality-specific artifacts affect downstream interpretation. 
Treating tools as interchangeable black-box operators, as is common in general-purpose agentic frameworks, obscures these modality-dependent constraints and can lead to physically implausible interpretations.

\subsection{Temporal Reasoning Beyond Sequential Modeling}

Temporal data in EO differ fundamentally from conventional sequential inputs such as video streams. 
Satellite observations are often irregularly sampled, interrupted by cloud cover, acquisition gaps, or sensor changes, and influenced by seasonal cycles and environmental processes. 
Many EO tasks, including change detection and impact assessment, require understanding the causes of observed change rather than merely identifying when change occurs~\cite{cheng2024change, fu2024remote}.  

This requirement extends beyond standard sequential modeling. 
Meaningful interpretation often involves distinguishing genuine environmental change from seasonal variability, sensor noise, or incomplete observations. 
Agentic reasoning strategies developed for regularly sampled environments therefore do not directly transfer. Effective EO reasoning must incorporate temporal context, periodic patterns, and domain knowledge to interpret changes reliably. Other EO pretraining and multimodal models likewise emphasize temporal structure through forecasting, irregular temporal sampling, seasonal contrast, and spatiotemporal prediction~\cite{smith2023earthpt, feng2025tessera, cong2022satmae, manas2021seasonal, yao2023ringmo}.

\subsection{Physical Constraints, Geospatial Validity, and Risk}

Remote sensing observations are governed by physical principles describing how sensors interact with the Earth's atmosphere and surface. 
Derived products, such as vegetation indices, land surface temperature estimates, or atmospheric variables, are meaningful only when computed under appropriate physical conditions and within established validity ranges~\cite{zhengming2002land, tucker1979red, NASA_MODIS}.  

Consequently, outputs that appear visually plausible may still be scientifically incorrect if underlying physical assumptions are violated or if earlier preprocessing steps introduce errors. This presents a challenge for agentic models that rely primarily on internal reasoning or self-evaluation signals. In EO workflows, correctness must ultimately be grounded in physical validity and geospatial consistency rather than internal coherence alone.

These constraints are further amplified by the high-impact nature of many remote sensing applications, including climate monitoring, disaster response, and environmental management. Errors introduced through incorrect tool orchestration, invalid preprocessing, or misinterpretation of intermediate outputs can propagate through analytical pipelines and influence downstream decisions with real-world consequences.

\begin{tcolorbox}[colback=gray!8, colframe=darkgold!75!black, boxrule=0.4pt, left=6pt, right=6pt, top=5pt, bottom=5pt]
{\large\color{darkgold!85!black}\faGlobe}\hspace{0.7em}%
\textbf{In EO, observation design is part of reasoning.} Scale, sensing modality, temporal window, and physical assumptions determine what can be measured, compared, and validly concluded downstream.
\end{tcolorbox}

Taken together, these structural properties do not merely complicate EO reasoning; they expose where generic agentic AI frameworks become fragile when transferred without adaptation.

\begin{figure}[t] 
  \centering
  \includegraphics[width=\columnwidth]{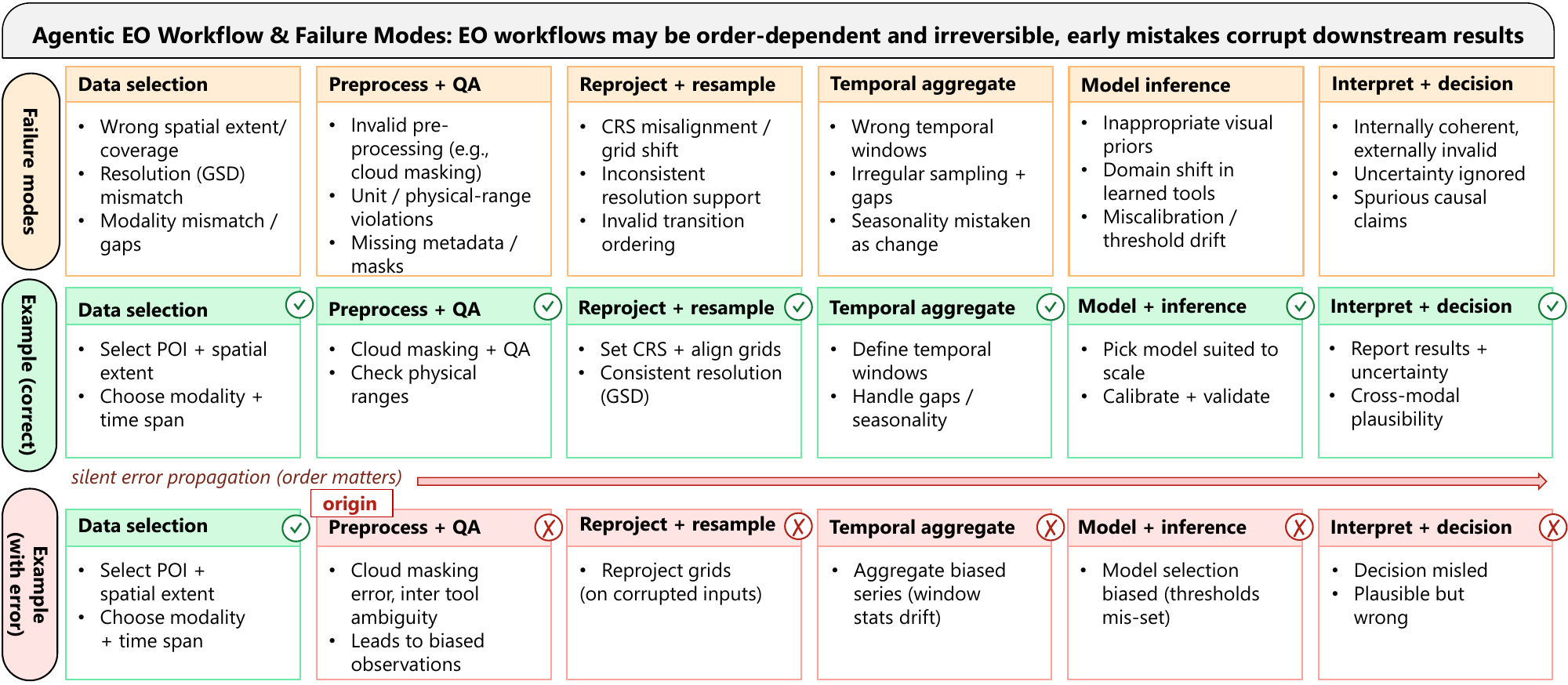}
  \caption{Agentic EO workflow and representative failure modes. The figure illustrates how errors introduced during data selection, preprocessing, reprojection, temporal aggregation, or model inference can propagate through downstream analytical stages. It also contrasts a geo-valid workflow with an error-prone workflow to highlight how early mistakes may silently corrupt later decisions.}
  \label{fig:eoFailures}
\end{figure}

\section{Limitations of Current Agentic AI in Remote Sensing}
\label{sec4}
The structural properties described in Section~\ref{sec3} expose several limitations in how current agentic AI models are designed, trained, and evaluated for EO. While recent agentic frameworks demonstrate strong performance in controlled or synthetic environments, their underlying assumptions do not always align with the realities of geospatial analytical workflows. When transferred to EO settings, these assumptions produce recurring failure modes in visual reasoning, tool orchestration, supervision, and long-horizon decision-making. This section examines those limitations and shows how generic agentic paradigms become structurally fragile in remote sensing. Representative examples of how such mismatches translate into downstream failures are shown in Figure~\ref{fig:eoFailures}.

\begin{tcolorbox}[colback=gray!8, colframe=darkgold!75!black, boxrule=0.4pt, left=6pt, right=6pt, top=5pt, bottom=5pt]
{\large\color{darkgold!85!black}\faGlobe}\hspace{0.7em}%
\textbf{Current agentic AI frameworks face systematic limits in EO.} Visual priors, black-box tool use, weak reward signals, and long-horizon reasoning become fragile under geospatial, physical, and operational constraints.
\end{tcolorbox}

\subsection{Transfer of Inappropriate Visual Priors}

Many agentic models inherit visual priors from natural-image pretraining or from vision-language models optimized for everyday scenes. These priors implicitly assume consistent object scales, well-defined visual boundaries, and relatively stable appearance statistics. In remote sensing, however, visual appearance varies substantially across sensors, spatial resolutions, viewing geometries, and environmental conditions. Consequently, features learned from natural imagery, or even from a single remote sensing modality, often generalize poorly across regions, spatial scales, or sensing configurations.

This limitation is visible in vision-centric agentic frameworks such as Chain of Visual Thoughts~\cite{qin2025chain}, OneThinker~\cite{feng2025onethinker}, and OpenThinkIMG~\cite{su2025openthinkimg}, where reasoning is largely grounded in stable visual semantics and object-level abstractions. In EO workflows, appearance-based reasoning can therefore produce interpretations that are internally plausible yet unstable under scale changes, modality shifts, or geographic transfer. More reliable EO reasoning must explicitly account for spatial context, measurement scale, and the physical processes underlying remote sensing observations.

\subsection{Tool Orchestration and Tool-Aware Optimization}

Tool orchestration is a central challenge for agentic models in remote sensing. Unlike the abstract tools used in many general-purpose benchmarks, geospatial tools are stateful, order-dependent, and often computationally intensive. Operations such as reprojection, resampling, mosaicking, spatial aggregation, and index computation must be executed in scientifically valid sequences and with appropriate parameterization to preserve geospatial validity. Once applied, many of these operations can irreversibly transform the underlying data, restricting later analytical options.

General orchestration frameworks such as ToolOrchestra~\cite{su2025toolorchestra}, OctoTools~\cite{lu2025octotools}, and VerlTool~\cite{jiang2025verltool} typically treat tools as interchangeable interfaces and focus on optimizing when and in what order they are invoked. While effective in domains with predictable tools and limited side effects, this abstraction obscures the semantics and persistent effects of geospatial transformations. In EO workflows, errors in tool choice, parameterization, or ordering may not cause immediate failure, but instead alter intermediate representations and propagate silently into later stages.

A related limitation is the weak treatment of tool-aware optimization. Many agentic models implicitly assume homogeneous action costs and optimize primarily for task success or execution efficiency. In EO, however, tool choices involve trade-offs among spatial coverage, resolution, computational cost, latency, and uncertainty accumulation. Recent work on tool-integrated reinforcement learning, including Tool Zero~\cite{zeng2025tool}, Tool-Augmented Policy Optimization~\cite{wu2025tool}, and VISTA-Gym~\cite{lu2025scaling}, shows that such trade-offs can be optimized in principle. Yet these methods typically assume simplified cost structures and rarely model domain-specific effects such as geospatial distortion, scale-dependent uncertainty, or irreversible preprocessing. Agents may therefore converge to workflows that are locally efficient yet scientifically fragile in real EO pipelines.

\subsection{Insufficient Supervision and Reward Signals}

Many agentic learning frameworks rely on text-level, task-level, or internally generated rewards to guide optimization. While such signals can be effective in domains where correctness can be verified through execution or internal consistency, they are often poorly aligned with the requirements of remote sensing. Textual correctness or task completion alone does not guarantee spatial accuracy, physical validity, or reliable uncertainty estimation.

This limitation becomes particularly visible in self-improving or self-evolving agentic models such as Agent0-VL~\cite{liu2025agent0}, EvoLMM~\cite{thawakar2025evolmm}, Multi-Agent Evolve~\cite{chen2025multi}, Vision-Zero~\cite{wang2025vision}, and AgentEvolver~\cite{zhai2025agentevolver}, which frequently derive learning signals from internal consistency, critique mechanisms, or judge-based evaluation. Recent analyses indicate that internally coherent reward loops may reinforce incorrect behaviors when external validity constraints are absent~\cite{shao2025your}. In EO, where correctness is determined by geospatial and physical constraints, the lack of externally grounded supervision remains a fundamental challenge.

\subsection{Long-Horizon Fragility}

Current agentic models also exhibit fragility over long decision horizons, a challenge that becomes particularly pronounced in EO workflows. Multi-step geospatial pipelines accumulate errors across preprocessing, analysis, and interpretation stages, often without explicit detection mechanisms. Memory components in existing agents are generally designed to maintain short-term reasoning traces rather than to track the validity of intermediate geospatial states across extended analytical chains.

Evaluation practices can further obscure these issues by focusing primarily on final outputs rather than the integrity of the analytical pipeline itself. This emphasis is visible in agentic benchmarks such as GAIA~\cite{mialon2023gaia}, VISTA-Gym~\cite{lu2025scaling}, and DeepEyes~\cite{zheng2025deepeyes}, where success is measured largely at the task level. In EO workflows, however, intermediate correctness is often critical: a model may produce a plausible final output while relying on invalid preprocessing steps or inconsistent spatial transformations. Addressing long-horizon fragility, therefore, requires evaluation protocols, memory mechanisms, and supervision strategies that explicitly account for pipeline integrity and error propagation. This concern is consistent with recent EO-agent evaluations in realistic tool-grounded environments, including GeoLLM-QA~\cite{singh2024evaluating}, GeoLLM-Engine~\cite{singh2024geollm}, Geo-OLM~\cite{stamoulis2025geo}, and UnivEARTH~\cite{kao2025towards}.

\subsection{When Agentic AI Should Not Be Used in EO}

The motivation for agentic AI in EO arises primarily in settings that require multi-step reasoning, adaptive decision-making, and interaction with heterogeneous geospatial tools. Not all EO workflows satisfy these conditions. Some tasks are fundamentally execution-oriented rather than decision-oriented, and treating them as standalone agentic problems can introduce unnecessary complexity without improving analytical capability.

Examples include atmospheric correction, radiometric calibration, reprojection to a predefined coordinate reference system, vegetation index computation, image tiling, mosaicking, and large-scale batch inference using trained detectors or segmentation models. These operations depend primarily on consistent execution rather than adaptive reasoning. This does not exclude them from agentic EO models, rather, they should function as reliable operators within broader workflows, where agentic intelligence emerges at the level of orchestration: deciding which sensing modalities to use, how spatial or temporal scope should be defined, and which analytical pathway best addresses a given objective.

Agentic deployment may also be premature when intermediate analytical states cannot be externally validated. Many current agentic paradigms rely on self-consistency, critique mechanisms, or learned evaluators to guide long-horizon optimization. When validation signals are weak, due to missing reference data, limited observability of preprocessing errors, or delayed manifestation of spatial inconsistencies, agents may converge to workflows that appear internally consistent while violating assumptions. In such cases, the limitation is not agentic AI alone, but the absence of the observability and validation infrastructure required for reliable agentic operation.

\section{Design Principles for Agentic Remote Sensing Models}
\label{sec5}

The limitations discussed in Section~\ref{sec4} suggest that adapting generic agentic AI paradigms to EO requires more than incremental changes to existing tool-use or reinforcement-learning frameworks. What is needed is a design perspective grounded in the structure of EO workflows themselves: geospatial data are stateful, transformations may be irreversible, correctness is externally defined, and intermediate analytical states matter. This section outlines design principles for agentic models that operate reliably over geospatial data, physically grounded tools, and structured EO state. The formulations below should be read as conceptual design patterns for EO-native agentic models rather than as a single prescribed implementation.

\begin{tcolorbox}[
  breakable,
  enhanced,
  colback=gray!6,
  colframe=black!60,
  boxrule=0.4pt,
  title=\textbf{Formalization: EO Agent Environment},
  before skip=8pt,
  after skip=8pt,
  top=4pt,
  bottom=4pt,
  left=6pt,
  right=6pt,
  boxsep=2pt
]
\setlength{\abovedisplayskip}{5pt}
\setlength{\belowdisplayskip}{5pt}
\setlength{\abovedisplayshortskip}{4pt}
\setlength{\belowdisplayshortskip}{4pt}

We model an EO agent as a constrained sequential decision-maker operating over structured geospatial state. At step $t$, the state is
\[
s_t = \left(x_t,\; c_t,\; r_t,\; e_t,\; \tau_t,\; m_t,\; u_t,\; p_t,\; h_t\right),
\]
where $x_t$ is the current data representation (e.g., an image tile, raster layer, mask, vector output, or derived product), $c_t$ the CRS, $r_t$ the spatial resolution or ground sampling distance (GSD), $e_t$ the spatial extent or area of support, $\tau_t$ the temporal window, $m_t$ the sensing modality, $u_t$ an uncertainty or reliability descriptor, $p_t$ provenance information describing how the current state was produced, and $h_t$ the tool-use history.

An action $a_t$ is a parameterized tool invocation,
\[
a_t = (\kappa_t,\; \theta_t),
\]
where $\kappa_t$ denotes the selected tool and $\theta_t$ its arguments. Executing $a_t$ induces a transition
\[
s_{t+1} \sim \mathcal{T}(\,\cdot \mid s_t, a_t\,),
\]
where the next state may reflect deterministic updates (e.g., clipping, reprojection, or metadata changes) as well as stochastic effects (e.g., cloud-masking ambiguity, interpolation artifacts, sensor noise, or learned-model uncertainty).
Valid EO transitions require both pre-execution feasibility and post-execution validity.
We represent these through action-level predicates $g(s_t,a_t)$, which determine whether a tool invocation is admissible in the current state, and transition-level verifier scores $V(s_t,a_t,s_{t+1})$, which evaluate the validity of the resulting state transition.
These checks include geometric consistency, temporal validity, radiometric/physical plausibility, provenance consistency, and statistical reliability.
\end{tcolorbox}

\begin{figure}[t] 
  \centering
  \includegraphics[width=0.9\columnwidth]{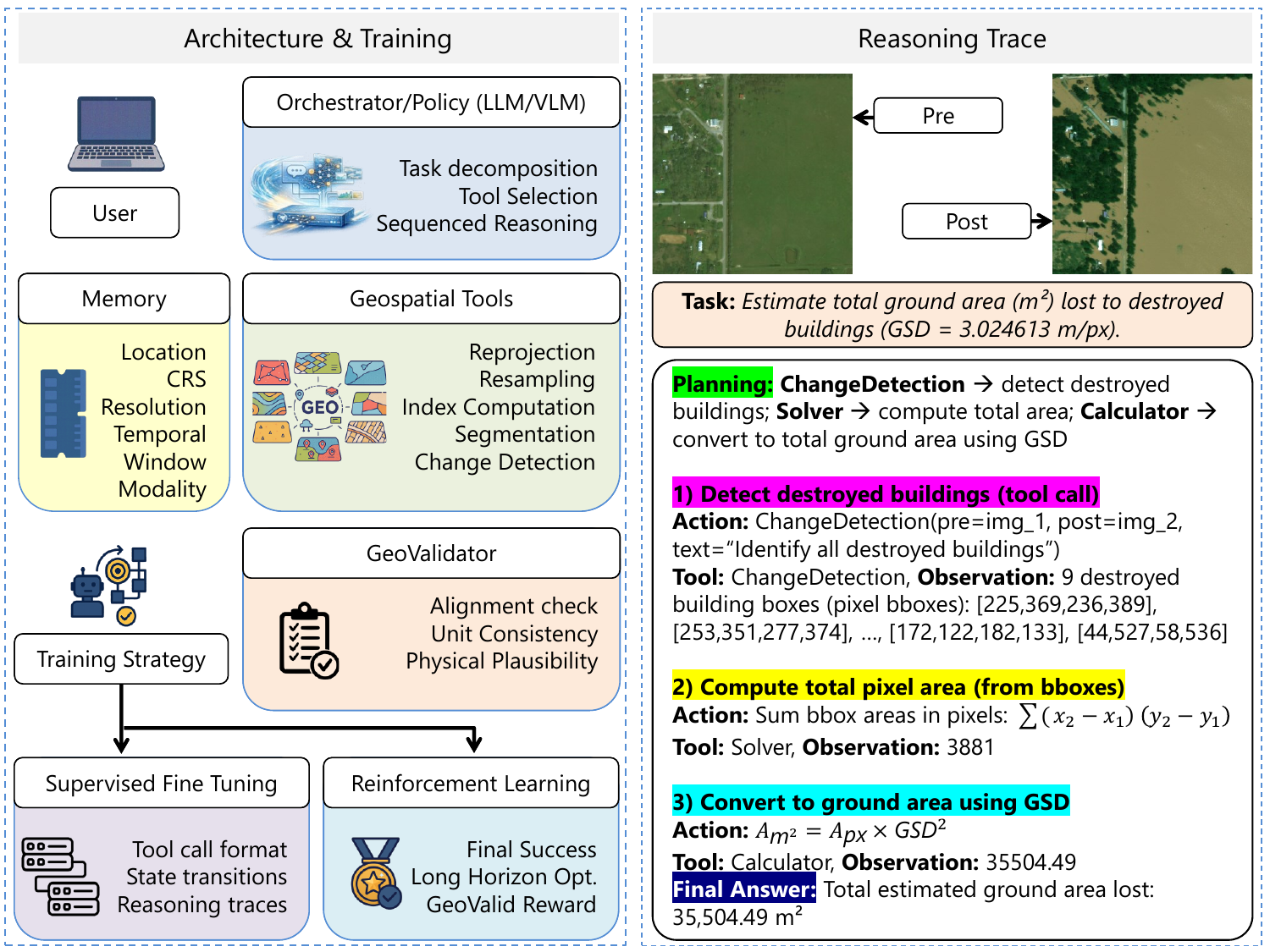}
  \caption{Agentic EO architecture and tool-integrated reasoning. \textbf{Left:} The agentic model includes an orchestrator policy (LLM/VLM), structured memory, geospatial tools, and a GeoValidator for enforcing spatial consistency. Training combines supervised fine-tuning and reinforcement learning with geospatial validity rewards. \textbf{Right:} Example reasoning trace showing sequential tool usage to estimate ground area lost to destroyed buildings from pre/post imagery using change detection and GSD-based area conversion. \textit{Drawings from ChatGPT.}}
  \label{fig:eoftrltrace}
\end{figure}

For notational simplicity, the subsequent formulations use the reduced state representation $s_t = (x_t, c_t, r_t, e_t, \tau_t, m_t, u_t)$, while provenance $p_t$ and tool history $h_t$ are treated as auxiliary metadata associated with the state. Figure~\ref{fig:eoftrltrace} provides a schematic view of this perspective, illustrating both the architectural components of an agentic EO model and a representative tool-integrated reasoning trace.

\subsection{Tool-Centered Agent Design}
\label{sec:tool_centered}

Building on the EO agent environment defined above, tool-centered design treats EO tools not as interchangeable APIs, but as operators that induce structured geospatial state transitions. In this setting, operations such as reprojection, resampling, spatial aggregation, spectral index computation, and temporal differencing are not merely auxiliary functions: they update the data, metadata, and validity conditions that govern subsequent reasoning. Because many such transformations are order-dependent and partly irreversible, tool choice and tool ordering become part of the analytical problem itself. Although early EO agents already rely on such semantics for model selection and geospatial reasoning~\cite{shabbir2025thinkgeo,chen2025remsa,hasan2025mapagent,feng2025earth}, most general tool-use frameworks~\cite{su2025toolorchestra,jiang2025verltool,wu2025tool} do not explicitly represent these constraints.

Using the reduced notation introduced above, we write the working geospatial state as
\begin{equation}
s_t = \left(x_t,\; c_t,\; r_t,\; e_t,\; \tau_t,\; m_t,\; u_t\right),
\end{equation}
where the variables retain the meanings introduced in the EO agent environment above. Although provenance $p_t$ and tool history $h_t$ are omitted from this compact notation, all feasibility and validation predicates below should be understood to have access to these auxiliary metadata when checks depend on provenance, reproducibility, or prior tool usage.
Tool execution then induces a conditional state transition
\begin{equation}
s_{t+1} \sim \mathcal{T}(\,\cdot \mid s_t, a_t),
\end{equation}
where $a_t$ denotes the selected tool invocation, including the tool identity, its arguments, and parameter values, and $\mathcal{T}(\,\cdot \mid s_t,a_t)$ denotes the transition model induced by applying that tool to the current EO state.
In practice, this means that each action modifies the analytical context over which later steps must operate. For example, reprojection changes the spatial reference frame, resampling changes effective resolution, clipping changes spatial support, and derived products such as change maps or spectral indices introduce new data representations.

Not every tool invocation is admissible for every state. Some actions are infeasible because the required inputs or state metadata are missing, regardless of how the policy is trained.
We therefore distinguish \emph{structural feasibility} from \emph{methodological quality}.
Structural feasibility asks whether a tool invocation is admissible at all from the current state, whereas methodological quality concerns how well a feasible transition preserves geospatial validity, physical validity, and reliability.
We first restrict the action space to structurally feasible tool invocations:
\begin{equation}
\mathcal{A}_{\mathrm{feas}}(s_t)
=
\left\{
a \in \mathcal{A}
\;:\;
g_{\mathrm{feas}}(s_t,a)=1
\right\},
\end{equation}
where $\mathcal{A}$ is the set of possible tool invocations and $g_{\mathrm{feas}}(s_t,a)$ is a hard feasibility predicate. One possible factorization is
\begin{equation}
g_{\mathrm{feas}}(s_t,a)
=
\mathbf{1}\!\left[
\phi_{\mathrm{crs}}(s_t,a)
\land
\phi_{\mathrm{ext}}(s_t,a)
\land
\phi_{\mathrm{res}}(s_t,a)
\land
\phi_{\mathrm{temp}}(s_t,a)
\land
\phi_{\mathrm{mod}}(s_t,a)
\land
\phi_{\mathrm{prov}}(s_t,a)
\right],
\end{equation}
where $\phi_{\mathrm{crs}}$, $\phi_{\mathrm{ext}}$, $\phi_{\mathrm{res}}$, $\phi_{\mathrm{temp}}$, $\phi_{\mathrm{mod}}$, and $\phi_{\mathrm{prov}}$ enforce the minimum preconditions required for meaningful execution, including valid coordinate-reference information, sufficient spatial extent or support, resolution compatibility, temporal support, modality availability, and provenance consistency. These checks prevent analytically inadmissible actions such as invoking a spectral index without the required bands, comparing states with missing temporal support, or producing outputs whose provenance is inconsistent with the workflow.

The central challenge of tool-centered EO agents is therefore not only deciding which tool to call, but learning policies over feasible state transitions in a geospatially constrained environment. Correspondingly, evaluation must examine not only final-answer correctness, but also whether the intermediate analytical pipeline remains scientifically coherent.

\subsection{Structured Reasoning Over Scale, Time, and Modality}
\label{sec:structured_reasoning}

Remote sensing reasoning is inherently multi-scale, multi-temporal, and multi-modal. Unlike natural-image reasoning, EO queries often require switching between localized high-resolution analysis, such as infrastructure damage assessment, and broader regional aggregation, such as flood extent or vegetation condition monitoring. Spatial scale determines which phenomena are observable and how uncertainty propagates. Temporal reasoning must account for irregular sampling, cloud contamination, sensor changes, and seasonal patterns rather than assuming regularly observed sequences. Multi-modal analysis, whether combining optical, SAR, multispectral, or reanalysis data, further introduces modality-specific uncertainty and complementary information content. Although existing EO-oriented agents demonstrate early forms of geospatial reasoning and querying~\cite{shabbir2025thinkgeo,hasan2025mapagent,feng2025earth,chen2025remsa}, most current agentic learning pipelines still reduce intermediate analytical states to unstructured text.

A structured approach instead requires explicit representation of at least four elements: (i) spatial extent and resolution, (ii) temporal windows and alignment constraints, (iii) modality selection and sensor-specific uncertainty, and (iv) confidence or uncertainty annotations attached to intermediate products. In practice, this means that change detection requires aligned grids and consistent radiometry, cross-modal analysis may require cloud masking or speckle-aware processing, and temporal comparisons must avoid leakage across inconsistent observation windows. 
Supervised trajectories should therefore encode state transitions such as extent changes, scale adjustments, and temporal selections rather than only listing tool names. Reinforcement objectives should similarly penalize scale misuse, spatial misalignment, and modality mismatch. A central challenge is how to jointly learn reasoning policies and structured EO state representations that remain stable across regions, sensors, and acquisition conditions.

\subsection{Tool Orchestration as Constrained Optimization}
\label{sec:orchestration_optimization}

In EO workflows, tool orchestration is better understood as a constrained, long-horizon optimization problem rather than a simple routing or sequencing task. Each operation modifies the underlying geospatial state and may restrict what can be inferred downstream. Transformations such as resampling, spatial aggregation, or filtering can partly irreversibly alter signal content, making early decisions particularly consequential. Moreover, the cost of tool execution depends strongly on spatial extent, spatial resolution, and sensing modality, introducing additional computational and resource constraints.

General orchestration and tool-reinforcement-learning frameworks~\cite{su2025toolorchestra,jiang2025verltool,wu2025tool,lu2025scaling} formalize tool selection and sequencing as part of the agent's policy. However, these approaches typically assume abstract cost models and semantics-free actions. In EO settings, orchestration must instead balance coupled objectives including analytical accuracy, spatial coverage, resolution fidelity, computational cost, latency, and uncertainty accumulation. The problem is therefore more naturally viewed as constrained trajectory optimization over reasoning and tool-use sequences.

Formally, let $\pi_{\theta}$ denote an agent policy parameterized by $\theta$, and let
$\tau = (s_0,a_0,s_1,a_1,\ldots,s_T)$ denote a reasoning-and-tool-use trajectory composed of EO states and tool invocations.
Because EO transitions may include stochastic effects, we write the trajectory distribution explicitly as $P(\tau \mid \pi_{\theta}, \mathcal{T})$, where $\mathcal{T}$ denotes the transition model introduced earlier.
Tool orchestration can then be expressed as the following constrained optimization problem:
\begin{equation}
\max_{\pi_{\theta}}
\;
\mathbb{E}_{\tau \sim P(\tau \mid \pi_{\theta}, \mathcal{T})}
\left[
\lambda_{\mathrm{ans}} R_{\mathrm{ans}}(s_T)
+
\sum_{t=0}^{T-1}\gamma^t\, r_{\mathrm{step}}(s_t,a_t,s_{t+1})
\right]
\quad
\text{s.t.}
\quad
g_{\mathrm{feas}}(s_t,a_t)=1,\;\forall t,
\qquad
C(\tau) \leq B,
\end{equation}
where $\gamma \in (0,1]$ is a discount factor, $R_{\mathrm{ans}}(s_T)$ denotes terminal task-level answer quality, $g_{\mathrm{feas}}(s_t,a_t)$ is the hard structural-feasibility constraint defined above, $C(\tau)$ denotes cumulative resource cost over the trajectory, and $B$ is a hard resource budget representing feasibility limits such as compute, latency, memory, or API usage.

We define the stepwise reward as
\begin{equation}
r_{\mathrm{step}}(s_t,a_t,s_{t+1})
=
\lambda_{\mathrm{geo}}\, q_{\mathrm{geo}}(s_t,a_t,s_{t+1})
+
\lambda_{\mathrm{phys}}\, r_{\mathrm{phys}}(s_t,a_t,s_{t+1})
-
\lambda_{\mathrm{cost}}\, c(s_t,a_t)
-
\lambda_{\mathrm{irr}}\, r_{\mathrm{irr}}(s_t,a_t,s_{t+1}),
\end{equation}
where $\lambda_{\mathrm{ans}},\lambda_{\mathrm{geo}},\lambda_{\mathrm{phys}},\lambda_{\mathrm{cost}},\lambda_{\mathrm{irr}} > 0$ are weighting coefficients. Here,
$q_{\mathrm{geo}}(s_t,a_t,s_{t+1}) \in [0,1]$ is a soft geospatial-quality score that evaluates the \emph{methodological quality} of a feasible transition,
$r_{\mathrm{phys}}(s_t,a_t,s_{t+1})$ captures physical validity,
$c(s_t,a_t)$ is the stepwise resource cost, and
$r_{\mathrm{irr}}(s_t,a_t,s_{t+1})$ penalizes irreversible operations that reduce downstream analytical flexibility.
In practice, concerns such as spatial coverage, resolution fidelity, and uncertainty accumulation can be encoded through the geospatial-quality and physical-validity terms, or through additional task-specific penalties when repeated transformations or low-reliability observations degrade downstream validity.

The hard feasibility predicate $g_{\mathrm{feas}}(s_t,a_t)$ and the soft geospatial-quality term $q_{\mathrm{geo}}(s_t,a_t,s_{t+1})$ serve different roles. The former asks whether a tool invocation is admissible at all in the current state, whereas the latter evaluates how well the chosen transition preserves geospatial quality among feasible alternatives. Likewise, the hard budget constraint $C(\tau)\leq B$ captures resource feasibility, while the penalty weighted by $\lambda_{\mathrm{cost}}$ encourages efficiency within the feasible set.

One possible definition of the soft geospatial-quality term is
\begin{equation}
q_{\mathrm{geo}}(s_t,a_t,s_{t+1})
=
\alpha_{\mathrm{crs}}\,\tilde v_{\mathrm{crs}}
+
\alpha_{\mathrm{align}}\,\tilde v_{\mathrm{align}}
+
\alpha_{\mathrm{temp}}\,\tilde v_{\mathrm{temp}}
+
\alpha_{\mathrm{unit}}\,\tilde v_{\mathrm{unit}}
+
\alpha_{\mathrm{prov}}\,\tilde v_{\mathrm{prov}},
\end{equation}
where $\tilde v_{\mathrm{crs}}, \tilde v_{\mathrm{align}}, \tilde v_{\mathrm{temp}}, \tilde v_{\mathrm{unit}}, \tilde v_{\mathrm{prov}} \in [0,1]$ are normalized sub-scores measuring, respectively, CRS consistency, spatial alignment, temporal validity, unit consistency, and provenance consistency of the transition $(s_t,a_t,s_{t+1})$, $\alpha_k \ge 0$, and $\sum_k \alpha_k = 1$.
This score allows feasible trajectories to be graded according to the geospatial quality of their intermediate transformations, rather than treating all feasible tool choices as equally good.

This formulation highlights an important distinction between EO orchestration and general tool-use optimization. In EO models, policy learning must account not only for terminal task success but also for structural feasibility of tool calls, methodological quality of intermediate transformations, physical validity, and resource efficiency. Consequently, evaluation should move beyond final success rates and instead assess trade-offs among analytical accuracy, geospatial validity, efficiency, and robustness of the resulting analytical pipeline. Tool-integrated reinforcement learning provides useful optimization machinery~\cite{wu2025tool,jiang2025verltool,lu2025scaling}, but EO-specific constraints must be made explicit through feasibility checks, graded quality terms, and scientifically grounded validation rather than treated as implicit side conditions.

\subsection{Learned Tools, Domain Shift, and Observation Reliability}
\label{sec:learned_tools}

In practical EO pipelines, many tools are learned models rather than deterministic scientific operators. These include segmentation networks for land-cover mapping, change detection models, super-resolution models, cross-modal translators, and region-specific classifiers. Unlike rule-based geospatial transformations such as reprojection or spectral index computation, learned tools encode statistical assumptions tied to their training distributions. When deployed across new geographic regions, spatial resolutions, sensing modalities, seasonal regimes, or acquisition geometries, their outputs may degrade systematically without producing explicit error signals.

Consequently, an agent may follow a geo-valid reasoning trace, with correct coordinate reference systems, aligned spatial grids, and consistent temporal windows, yet still obtain semantically incorrect intermediate results because the underlying learned model is miscalibrated or operating outside its training distribution. Observation reliability under domain shift therefore becomes a bottleneck that is independent of tool orchestration quality.

This limitation cannot be addressed solely through improved planning or tool sequencing. Agentic EO models must also reason explicitly about the reliability of tool outputs and, when necessary, adapt learned tools to the deployment environment. One way to capture this is to associate each learned-tool output with a reliability estimate conditioned on the current EO state:
\begin{equation}
q_t = Q_{\psi}(z_t, s_t),
\end{equation}
where $z_t$ denotes the output produced by a learned tool at step $t$, $s_t$ represents the current structured EO state defined in Section~\ref{sec:tool_centered}, and $q_t \in [0,1]$ is a reliability score estimating the likelihood that the tool output is valid under the current spatial, temporal, and modality conditions.
In practice, the learned reliability estimate $q_t$ can be used to instantiate or update the uncertainty component of the next EO state, i.e., $u_{t+1}$, so that downstream decisions are conditioned not only on the tool output itself but also on how trustworthy that output is under the current deployment conditions.

A related concept is the deployment shift between training and operational conditions. This shift can be expressed as the discrepancy between the distribution of data used to train the model and the distribution encountered during deployment:
\begin{equation}
\Delta_{\mathrm{shift}}
=
d\!\left(
p_{\mathrm{train}}(x,m,r,\tau),\;
p_{\mathrm{test}}(x,m,r,\tau)
\right),
\end{equation}
where $p_{\mathrm{train}}$ and $p_{\mathrm{test}}$ denote the joint distributions of observations $x$, sensing modality $m$, spatial resolution $r$, and temporal regime $\tau$ in the training and deployment environments, respectively, and $d(\cdot,\cdot)$ measures the discrepancy between these distributions (for example using divergence or distance metrics).
Large values of $\Delta_{\mathrm{shift}}$ can be used as contextual evidence that the reliability of learned-tool outputs may degrade under deployment, and may therefore inform the reliability estimate.

In practice, addressing such shifts may require lightweight adaptation of learned tools conditioned on region, modality, or acquisition regime. For instance, a segmentation model may be recalibrated using region-specific statistics or adapted using a small number of labeled tiles before being deployed within a broader analytical workflow. Adaptation may take the form of parameter-efficient tuning, threshold re-estimation, calibration, or domain-specific normalization. 
Importantly, this adaptation step should itself become part of the agent's decision space: a planner may determine whether adaptation is necessary, while a verifier may assess whether post-adaptation outputs satisfy geospatial, temporal, and modality consistency constraints. 
Without explicit modeling of tool reliability and deployment shift, agentic optimization risks amplifying systematic biases and producing outputs that are structurally valid yet scientifically misleading.

\subsection{Complementary Roles of Supervised and Reinforcement Learning}
\label{sec:sft_rl_roles}

Supervised fine-tuning (SFT) and reinforcement learning (RL) play complementary roles in the design of agentic EO models. 
SFT is particularly effective for stabilizing structured interaction protocols, including formatting geo-queries, selecting spatial extents, parameterizing reprojection and resampling operations, handling masks, and generating correctly formatted tool invocations. 
In many tool-augmented reasoning pipelines, supervision is applied to reasoning traces and tool invocation sequences~\cite{jiang2025verltool,lu2025scaling,su2025openthinkimg}, enabling the model to learn when and how tools should be called.

For clarity, we write the supervised objective over model-generated reasoning and action outputs, while treating tool outputs as observations returned by the environment and appended to the agent context.
This matches many policy-learning formulations and preserves modularity between the reasoning policy and external operators.
Some cold-start pipelines also supervise full action-observation trajectories~\cite{su2025openthinkimg}.
In tool-integrated RL settings, by contrast, observation tokens are often treated as off-policy and masked during policy optimization~\cite{jiang2025verltool}.

Let $\tau = (s_0,a_0,s_1,a_1,\ldots,s_T)$ denote a reasoning-and-tool-use trajectory composed of EO states and tool invocations. 
Under supervised fine-tuning, the policy can be trained over such trajectories using a standard sequence likelihood objective:
\begin{equation}
\mathcal{L}_{\mathrm{SFT}}
=
-
\sum_{i=1}^{N}
\sum_{t=1}^{T_i}
\log \pi_{\theta}
\left(
y^{(i)}_t
\mid
y^{(i)}_{<t},
o^{(i)},
s^{(i)}_t
\right),
\end{equation}
where $o^{(i)}$ denotes the observation context available for training example $i$, 
$s^{(i)}_t$ is the structured EO state at step $t$, and 
$y^{(i)}_t$ represents the next model-generated reasoning or action unit at step $t$, including tool identifiers and structured arguments such as CRS parameters or spatial extents.

While SFT stabilizes reasoning structure and tool-call syntax, reinforcement learning is better suited for optimizing long-horizon decision making. 
In EO workflows, decisions such as selecting spatial scale, switching sensing modalities, allocating compute across large spatial extents, or acquiring additional temporal observations often have delayed consequences that cannot be easily captured through supervised labels.

A commonly used policy optimization objective is the clipped surrogate objective:
\begin{equation}
\mathcal{L}_{\mathrm{RL}}(\theta)
=
-
\mathbb{E}_{t}
\left[
\min\!\left(
\rho_t A_t,
\mathrm{clip}(\rho_t,1-\epsilon,1+\epsilon)A_t
\right)
\right],
\qquad
\rho_t
=
\frac{\pi_{\theta}(a_t \mid s_t)}{\pi_{\theta_{\mathrm{old}}}(a_t \mid s_t)},
\end{equation}
where $\mathcal{L}_{\mathrm{RL}}$ is written as a loss to be minimized, $\rho_t$ is the policy ratio between the current and previous policies, and $A_t$ is an advantage estimate associated with the selected action.

To make delayed EO consequences explicit, let $r_t$ denote the step-level reward obtained after executing action $a_t$ in state $s_t$. The discounted return from time step $t$ is
\begin{equation}
G_t^{\gamma}
=
\sum_{k=0}^{T-1-t}
\gamma^{k} r_{t+k},
\end{equation}
where $\gamma \in (0,1]$ controls how strongly future rewards influence the current decision. Smaller values of $\gamma$ emphasize near-term rewards, whereas values closer to one assign greater importance to downstream consequences such as delayed geospatial misalignment, accumulated uncertainty, or final-answer correctness.

In practice, $A_t$ can be estimated from discounted returns, for example by centering or normalizing $G_t^{\gamma}$ within a batch, or by subtracting a learned baseline. The precise estimator is implementation-dependent, the key point for EO is that the learning signal should reflect trajectory-level consequences rather than only immediate tool success.

In EO settings, the step reward $r_t$ may incorporate components reflecting task correctness, geospatial validity checks, physical validity constraints, and computational cost penalties associated with the analytical pipeline. 
Designing reward signals that capture these properties while remaining automatically verifiable remains a central challenge for scalable agentic EO models.

\subsection{Multi-Agent Decomposition as Geospatial Workflow}
\label{sec:multiagent_workflow}

\begin{figure}[t] 
  \centering
  \includegraphics[width=0.9\columnwidth]{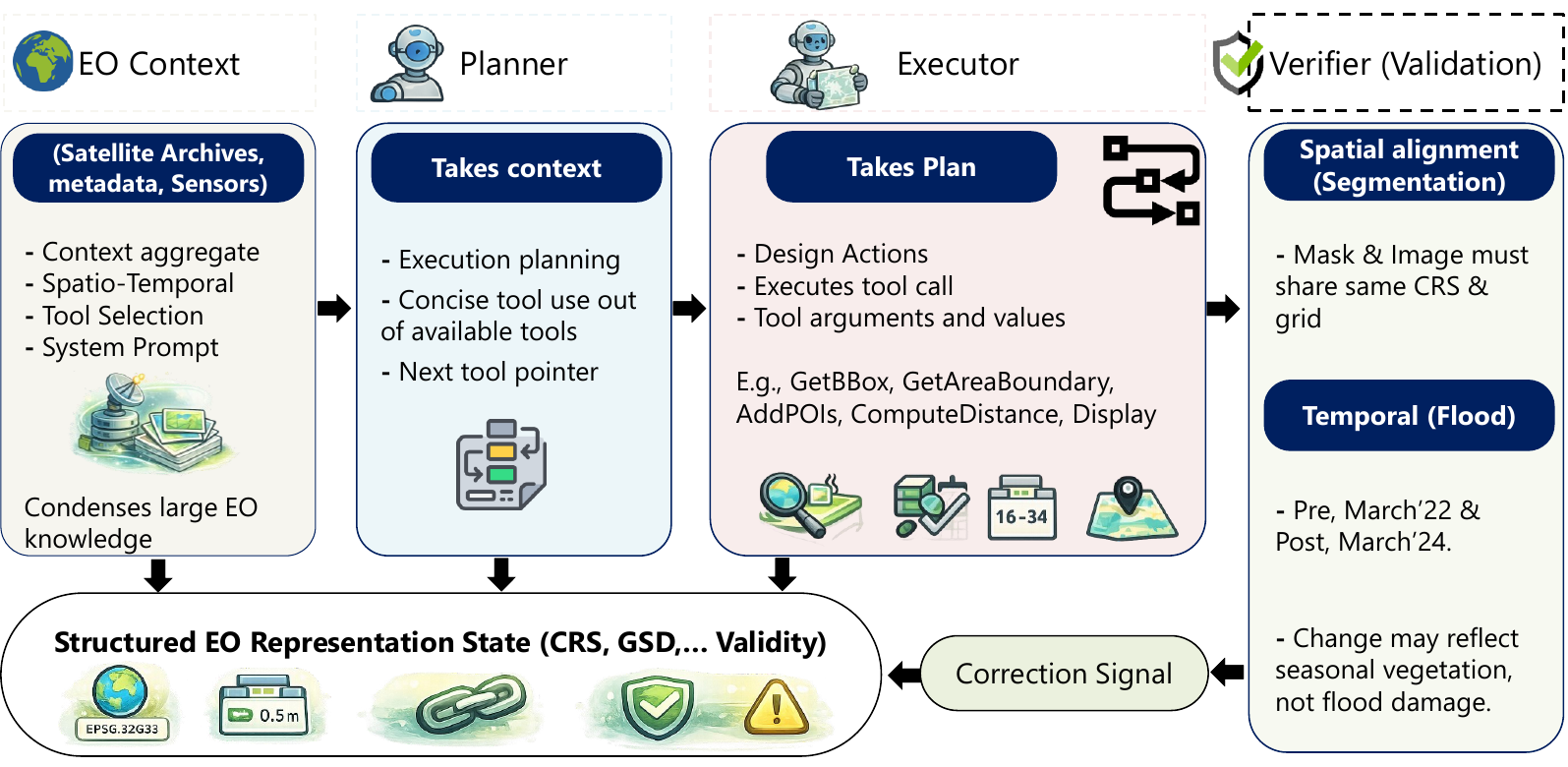}
  \caption{Design blueprint for agentic Earth observation. This consists of a Planner, Executor, and Verifier operating over a shared structured EO state representation. The Planner condenses large EO context into an execution plan, the Executor performs tool calls with explicit arguments, and the Verifier ensures geospatial and physical validity through checks such as CRS consistency, spatial alignment, temporal validity, and physically meaningful interpretation (e.g., distinguishing seasonal vegetation change from flood damage). \textit{Drawings from ChatGPT.}}
  \label{fig:eopev}
\end{figure}

Multi-agent decomposition in EO models should mirror the structure of geospatial analysis rather than generic modular engineering patterns. In practical geospatial workflows, different stages of analysis involve distinct reasoning tasks and validation requirements. A corresponding planner-executor-verifier abstraction is illustrated in Figure~\ref{fig:eopev}, which highlights the role of structured EO state and explicit geospatial validation in multi-agent EO reasoning. A planner agent should formulate analytical hypotheses, determine the spatial and temporal scope of the investigation, and select appropriate sensing modalities. An executor agent should then invoke geospatial tools with correct parameters, such as CRS, spatial resolution, spatial extent, and temporal windows, while maintaining the sequence of intermediate analytical states.
Finally, a verifier agent should perform domain-grounded validation of intermediate results, including spatial alignment audits, unit consistency checks, temporal-window validity checks, and physical validity checks on derived quantities.

Although extensible agent frameworks~\cite{lu2025octotools,su2025toolorchestra} enable modular tool routing and workflow composition, and EO-specific models such as MapAgent~\cite{hasan2025mapagent} demonstrate hierarchical task decomposition, most existing agentic models do not explicitly formalize verifier roles grounded in geospatial validity or enforce structured validation of intermediate states.
As a result, an agent may produce internally coherent reasoning traces while still violating geospatial constraints or physical assumptions embedded in EO data.

This decomposition also enables specialization across scale reasoning, temporal alignment, modality selection, and uncertainty assessment, while allowing verification to remain targeted at different stages of the workflow.

A verifier agent can therefore evaluate the validity of each state transition induced by tool execution. Rather than relying on a single scalar check, the verifier can be organized as a small taxonomy of complementary audits over geometry, time, physical validity, provenance, and statistical reliability. Given a state transition $(s_t, a_t, s_{t+1})$, the verifier score can be written as
\begin{equation}
V(s_t, a_t, s_{t+1})
=
\lambda_{\mathrm{geom}} v_{\mathrm{geom}}
+
\lambda_{\mathrm{temp}} v_{\mathrm{temp}}
+
\lambda_{\mathrm{phys}} v_{\mathrm{phys}}
+
\lambda_{\mathrm{prov}} v_{\mathrm{prov}}
+
\lambda_{\mathrm{stat}} v_{\mathrm{stat}},
\end{equation}
where $\lambda_{\mathrm{geom}}, \lambda_{\mathrm{temp}}, \lambda_{\mathrm{phys}}, \lambda_{\mathrm{prov}}, \lambda_{\mathrm{stat}} > 0$, and $v_{\mathrm{geom}}, v_{\mathrm{temp}}, v_{\mathrm{phys}}, v_{\mathrm{prov}}, v_{\mathrm{stat}} \in [0,1]$:
\begin{itemize}
    \item \textbf{Geometric verifiers} $v_{\mathrm{geom}}$ check whether the transition is spatially coherent, including CRS compatibility, reprojection sanity, raster alignment, and pixel-grid or resolution consistency.
    
    \item \textbf{Temporal verifiers} $v_{\mathrm{temp}}$ check whether the selected observations are temporally appropriate, including window overlap, absence of temporal leakage, and seasonal comparability when pre/post comparisons are made.
    
    \item \textbf{Radiometric and physical verifiers} $v_{\mathrm{phys}}$ check whether intermediate outputs remain physically valid, including valid reflectance or index ranges, physically meaningful derived quantities, and consistency of measurement units.
    
    \item \textbf{Provenance verifiers} $v_{\mathrm{prov}}$ check whether intermediate products remain traceable and reproducible, including source-data provenance, tool versioning, parameter logging, and consistency of the recorded transformation history.
    
    \item \textbf{Statistical verifiers} $v_{\mathrm{stat}}$ check whether outputs remain reliable under uncertainty, including uncertainty calibration, low-confidence warnings, and simple distribution-shift or anomaly alarms when tool outputs are inconsistent with expected operating conditions.
\end{itemize}

For interpretability, the verifier score can be normalized as
\begin{equation}
\tilde{V}(s_t,a_t,s_{t+1})
=
\frac{V(s_t,a_t,s_{t+1})}{\lambda_{\mathrm{geom}}+\lambda_{\mathrm{temp}}+\lambda_{\mathrm{phys}}+\lambda_{\mathrm{prov}}+\lambda_{\mathrm{stat}}},
\end{equation}
yielding a normalized verification score in the range $[0,1]$.

This formulation highlights the role of verifier agents in EO pipelines. Unlike language-model self-critique, which primarily evaluates logical consistency, verifier agents must combine deterministic geospatial audits with probabilistic checks based on uncertainty propagation or cross-modality corroboration. Without such domain-grounded verification loops, multi-agent models may reinforce internally coherent yet scientifically invalid reasoning chains. Future EO agent architectures should therefore integrate role specialization with explicit geo-valid state auditing, so that decomposition improves analytical reliability rather than merely distributing reasoning complexity.

\section{Research Directions}
\label{sec6}

The design principles outlined in Section~\ref{sec5} suggest that reliable agentic EO models will require advances not only in model architectures but also in evaluation resources, training strategies, and validation protocols. Progress therefore depends on coordinated development across datasets, learning paradigms, self-improvement mechanisms, and evaluation frameworks that reflect the structure of EO workflows. This section outlines research directions for developing scalable, geo-valid, and physically grounded agentic remote-sensing models.

\begin{tcolorbox}[colback=gray!8, colframe=darkgold!75!black, boxrule=0.4pt, left=6pt, right=6pt, top=5pt, bottom=5pt]
{\large\color{darkgold!85!black}\faGlobe}\hspace{0.7em}%
\textbf{Advancing agentic EO requires infrastructure for valid reasoning.} Benchmarks, learning objectives, self-improvement loops, and evaluation protocols must encode geospatial state, tool effects, and external validity.
\end{tcolorbox}

\subsection{Datasets and Benchmarks}

The absence of domain-specific evaluation resources remains a major bottleneck for developing agentic remote sensing models. Most existing agentic benchmarks assume generic action spaces, treat tools as black-box APIs, and emphasize terminal task success, underrepresenting the multi-stage nature of EO workflows. In practical RS applications, tasks frequently require coordinating optical and SAR imagery, querying temporal records from multispectral or reanalysis datasets, and executing ordered transformations such as reprojection, cloud masking, spectral index computation, spatial aggregation, and change detection. Each stage introduces constraints on spatial extent, CRS, spatial resolution or GSD, and temporal alignment.

To be useful, an EO-agent benchmark should specify the task families the agent is expected to solve, the tool set available for interaction, the success criteria used for evaluation, and the data split protocol used to prevent spatial or temporal leakage. It should also make clear whether intermediate tool traces or expert-validated state transitions are available, and whether operational constraints such as tool-call budgets, latency, or compute cost are part of the evaluation. Without these design choices, benchmark results are difficult to interpret: strong performance may reflect shortcut exploitation, leakage across regions or seasons, or dependence on tools that are unrealistically restricted or overly permissive.

Future benchmarks should extend this realism through additional multi-sensor sources, irregular time series, realistic tool constraints, and expert-verified intermediate traces. Evaluation should combine terminal answer quality with trajectory-level criteria such as pipeline integrity, geospatial validity, and efficiency, so that scientifically weak but superficially plausible workflows are not rewarded. Related resources moving toward more realistic evaluation include GeoLLM-Engine~\cite{singh2024geollm}, XLRS-Bench~\cite{wang2025xlrs}, ThinkGeo~\cite{shabbir2025thinkgeo} and OpenEarthAgent~\cite{shabbir2026openearthagent}.

\subsection{Hybrid Learning Strategies}

A central research question is how to train EO agents that combine the structural reliability of supervised trajectories with the long-horizon adaptability of reinforcement learning. While Section~\ref{sec:sft_rl_roles} described the complementary roles of SFT and RL, several open problems remain in making this combination effective under geospatial constraints. In particular, future work must determine how structured EO state should be represented during training, how expert demonstrations should encode valid transitions over spatial extent, resolution, time, and modality, and how such supervision can generalize across regions, sensors, and tasks.

A second challenge concerns long-horizon policy optimization under delayed and partially observable consequences. In EO workflows, decisions about scale, modality, temporal retrieval, and compute allocation can affect downstream validity in ways that are difficult to supervise directly. This raises open questions about reward design, curriculum construction, and policy transfer: which components of geospatial validity can be checked automatically, how geo-valid and physically meaningful rewards should be balanced against efficiency, and how policies can remain robust under domain shift across locations, seasons, and sensing conditions.
Promising directions include supervised trajectory learning combined with constrained RL refinement, curricula that gradually increase spatial-temporal complexity, and transfer strategies that improve generalization across heterogeneous EO environments~\cite{su2025openthinkimg,jiang2025verltool,wu2025tool}.

At a high level, these components can be combined through a hybrid training objective
\begin{equation}
\mathcal{L}_{\mathrm{hybrid}}
=
\mathcal{L}_{\mathrm{SFT}}
+
\eta\,\mathcal{L}_{\mathrm{RL}},
\end{equation}
where $\mathcal{L}_{\mathrm{SFT}}$ represents the supervised trajectory imitation objective defined earlier, $\mathcal{L}_{\mathrm{RL}}$ denotes the policy optimization loss over reasoning and tool-use trajectories, and $\eta \ge 0$ controls the relative contribution of long-horizon reinforcement learning during training.

\subsection{Constrained Self-Improvement}

A key research direction is how to make self-improving EO agents scalable without allowing them to drift toward internally consistent but scientifically invalid behavior. In EO settings, internal coherence does not guarantee physical validity: an agent may generate plausible land-cover maps that violate radiometric constraints, compute derived quantities outside meaningful ranges, or reinforce invalid transformation sequences in multi-step pipelines.

For EO applications, self-improvement mechanisms must therefore remain constrained by externally verifiable conditions. Deterministic validation modules can enforce fundamental geospatial constraints such as CRS compatibility, grid alignment, valid spatial transformations, and temporally consistent observation windows. Additional checks may verify that derived quantities fall within physically meaningful ranges (e.g., vegetation indices or reflectance values) and that transformations preserve spatial consistency. Complementary probabilistic checks may assess uncertainty propagation, sensor noise characteristics, or cross-modal agreement, for example when SAR observations corroborate optical measurements under cloud cover.

Self-generated tasks should remain grounded in realistic EO scenarios and preserve geospatial realism. Synthetic augmentations must respect radiative, atmospheric, and geometric constraints that govern sensor measurements. Moreover, all automatically generated analytical workflows should remain auditable. Tool parameters, input data sources, and intermediate state transitions must be recorded so that analytical decisions can be inspected and reproduced.

These requirements can be expressed as a constrained self-improvement objective over reasoning and tool-use trajectories:
\begin{equation}
\max_{\pi_{\theta}}
\;
\mathbb{E}_{\tau \sim P(\tau \mid \pi_{\theta}, \mathcal{T})}
\left[
R_{\mathrm{self}}(\tau)
\right]
\quad
\text{s.t.}
\quad
\tilde{V}(s_t,a_t,s_{t+1}) \ge \delta,
\;\forall t,
\end{equation}
where $\pi_{\theta}$ denotes the agent policy, $\tau$ is a reasoning-and-tool-use trajectory, $R_{\mathrm{self}}(\tau)$ represents internally generated improvement signals, and $\tilde{V}(s_t,a_t,s_{t+1})$ denotes the normalized verifier score over the induced EO state transition. The threshold $\delta \in [0,1]$ specifies the minimum level of transition validity required for self-generated trajectories to be accepted during training.

To ensure reproducibility and traceability of automatically generated analyses, each trajectory should also maintain an execution record:
\begin{equation}
\mathcal{H}(\tau)
=
\big((\kappa_t,\; \theta_t,\; d_t,\; s_t,\; s_{t+1})\big)_{t=0}^{T-1},
\end{equation}
where $\kappa_t$ denotes the selected tool, $\theta_t$ its parameters (e.g., CRS specification, spatial extent, or resampling method), $d_t$ denotes the data sources used at step $t$, and $(s_t,s_{t+1})$ records the pre and post action EO states associated with that transition.

Embedding these validation constraints within self-improving learning loops is essential to prevent agents from drifting toward internally consistent but externally invalid reasoning strategies. By enforcing verifier-based acceptance criteria and maintaining analytical traces, self-evolving EO agents can improve while preserving reliability and reproducibility.

\subsection{Evaluation Beyond Accuracy}

Evaluating agentic remote sensing models solely by final-answer accuracy is insufficient. In EO workflows, analytical outputs are produced through multi-step reasoning and tool-execution pipelines. A model may therefore produce a correct final answer while relying on invalid intermediate operations, such as misaligned CRS, inconsistent spatial resolution, incorrect temporal windows, or improper unit conversions. Such errors may propagate through the analytical pipeline and compromise the validity of the result.

Evaluation should therefore extend from single outputs to the full reasoning trajectory. 
Let $\tau = (s_0,a_0,s_1,\ldots,a_{T-1},s_T)$ denote a trajectory of intermediate EO states and tool actions.
Beyond final-task accuracy, several trajectory-level criteria can be defined.

\paragraph{Pipeline Integrity.}
A fundamental requirement is that each analytical step remains valid. This can be measured as the fraction of steps that satisfy verifier-based constraints:
\begin{equation}
\mathrm{PI}(\tau)
=
\frac{1}{T}
\sum_{t=0}^{T-1}
\mathbf{1}\!\left[
\tilde{V}(s_t,a_t,s_{t+1}) \ge \delta
\right],
\end{equation}
where $\tilde{V}(s_t,a_t,s_{t+1})$ denotes the normalized verifier score associated with the transition from $s_t$ to $s_{t+1}$ and $\delta \in [0,1]$ represents the minimum acceptable transition-validity threshold.

\paragraph{Trajectory Validity Score.}
A softer measure evaluates the average validity of the trajectory by averaging the normalized verifier scores across steps:
\begin{equation}
\mathrm{TVS}(\tau)
=
\frac{1}{T}
\sum_{t=0}^{T-1}
\tilde{V}(s_t,a_t,s_{t+1}),
\end{equation}
where $\tilde{V}$ denotes the normalized verifier score defined in Section~\ref{sec:multiagent_workflow}.

\paragraph{Discounted Inconsistency Burden.}
To summarize trajectory-level inconsistency while emphasizing earlier errors, one can define a discounted inconsistency burden:
\begin{equation}
\mathrm{DIB}(\tau)
=
\frac{\sum_{t=0}^{T-1} \beta^{t}\,\ell_t}
{\sum_{t=0}^{T-1} \beta^{t}},
\end{equation}
where $\ell_t$ denotes the inconsistency detected at step $t$, and $\beta \in (0,1]$ is a weighting factor that assigns greater weight to earlier inconsistencies when $\beta<1$.

\paragraph{Cost-Aware Efficiency.}
Because EO analyses often operate over large spatial extents and high-resolution imagery, evaluation should also consider computational efficiency:
\begin{equation}
\mathrm{Eff}(\tau)
=
\frac{R_{\mathrm{ans}}(\tau)}{1 + C(\tau)},
\end{equation}
where $R_{\mathrm{ans}}(\tau) \in [0,1]$ measures normalized final-task answer quality and $C(\tau) \ge 0$ denotes a normalized or dimensionless trajectory cost, such as normalized runtime, tool-call budget, memory use, or API cost.

Together, these trajectory-level criteria move evaluation beyond final-answer correctness toward assessing whether agentic EO models operate through scientifically valid, geospatially grounded, and computationally efficient analytical pipelines. More broadly, progress in agentic EO will depend not only on advances in model capability, but also on benchmarks, learning strategies, and evaluation protocols that reflect the scientific structure of geospatial workflows. The central challenge is therefore not merely to improve task performance, but to develop agents whose reasoning remains geo-valid, physically grounded, and robust over long-horizon analysis.

\begin{tcolorbox}[colback=gray!8, colframe=darkgold!75!black, boxrule=0.4pt, left=6pt, right=6pt, top=5pt, bottom=5pt]
{\large\color{darkgold!85!black}\faGlobe}\hspace{0.7em}%
\textbf{In EO, evaluation must move from answer correctness to pipeline validity.} What matters is whether the full trajectory remains geo-valid, physically grounded, reproducible, and efficient.
\end{tcolorbox}

\section{Conclusion}
\label{sec7}
Agentic AI offers a promising framework for analytical problems that require multi-step reasoning, adaptive decision making, and interaction with heterogeneous tools. However, EO workflows introduce constraints that are largely absent from domains of current agentic models. Remote sensing data are georeferenced, multi-modal, and temporally indexed, and analytical pipelines often involve stateful transformations whose validity depends on spatial reference, resolution consistency, and temporal alignment. As a result, naively transferring generic agentic paradigms to EO can produce workflows that are internally coherent yet scientifically invalid. This paper argues that reliable agentic EO models must explicitly represent structured geospatial states, treat tools as domain-constrained operators, and evaluate reasoning not only by final answers but also by the geospatial validity of intermediate analytical steps.
Looking ahead, agentic remote sensing models should be developed as decision-support instruments embedded within operational workflows rather than as standalone predictive models. Their value lies not in replacing established EO methods, but in coordinating data, tools, and domain knowledge under geospatial and physical constraints. Realizing this role will require not only advances in model capability, but also transparent reasoning traces, stronger geospatial validation, and carefully designed interfaces among agents, tools, and human analysts.

\bibliographystyle{ACM-Reference-Format}
\bibliography{sample-base}

\end{document}